\documentclass[10pt,twocolumn,letterpaper]{article}

\usepackage[pagenumbers]{cvpr} 
\usepackage{comment}

\usepackage{graphicx}
\usepackage{amsmath}
\usepackage{amssymb}
\usepackage{booktabs}

\usepackage[dvipsnames]{xcolor}

\usepackage{times}
\usepackage{epsfig}
\usepackage{graphicx}
\usepackage{amsmath}
\usepackage{amssymb}
\usepackage{multirow}
\usepackage{diagbox}
\usepackage{floatrow} 
\usepackage[font=small]{caption}
\usepackage{float}
\usepackage{dblfloatfix}
\usepackage{setspace}
\usepackage[accsupp]{axessibility}

\definecolor{cvprblue}{rgb}{0.21,0.49,0.74}
\usepackage[pagebackref,breaklinks,colorlinks,citecolor=cvprblue]{hyperref}

\def\paperID{2119} 
\def\confName{CVPR}
\def\confYear{2024}

\begin{document}
\title{SANeRF-HQ: Segment Anything for NeRF in High Quality}


\author{
\begin{tabular}{cccc}
Yichen Liu$^{1}$ & 
Benran Hu$^{2}$ & 
Chi-Keung Tang$^{1}$ &
Yu-Wing Tai$^{3}$ 
\end{tabular}
\\
\begin{tabular}{cc}
\multicolumn{2}{c}{$^1$The Hong Kong University of Science and Technology} \\
$^2$Carnegie Mellon University &
$^3$Dartmouth College  
\end{tabular} \\
}

\maketitle
\renewcommand{\thefootnote}{\fnsymbol{footnote}}

\begin{abstract}

\vspace{-0.104in}
Recently, the Segment Anything Model (SAM) has showcased remarkable capabilities of zero-shot segmentation, while  NeRF (Neural Radiance Fields) has gained popularity as a method for various 3D problems beyond novel view synthesis. Though there exist initial attempts to incorporate these two methods into 3D segmentation, they face the challenge of accurately and consistently segmenting objects in complex scenarios. In this paper, we introduce the Segment Anything for NeRF in High Quality (SANeRF-HQ) to achieve high-quality 3D segmentation of any target object in a given scene. SANeRF-HQ utilizes SAM for open-world object segmentation guided by user-supplied prompts, while leveraging NeRF to aggregate information from different viewpoints. To overcome the aforementioned challenges, we employ density field and RGB similarity to enhance the accuracy of segmentation boundary during the aggregation. Emphasizing on segmentation accuracy, we evaluate our method on multiple NeRF datasets where high-quality ground-truths are available or manually annotated. SANeRF-HQ shows a significant quality improvement over state-of-the-art methods in NeRF object segmentation, provides higher flexibility for object localization, and enables more consistent object segmentation across multiple views. Results and code are available at the project site: \href{https://lyclyc52.github.io/SANeRF-HQ/}{https://lyclyc52.github.io/SANeRF-HQ/}.

\end{abstract}  
\footnotetext[2]{This research is supported in part by the Research Grant Council of the Hong Kong SAR under grant no. 16201420.}
\vspace{-0.2in}
\section{Introduction}
\label{sec:intro}

Neural Radiance Field (NeRF)~\cite{mildenhall2020nerf} has produced state-of-the-art results in novel view synthesis for intricate real-world scenes. NeRF encodes a given scene using Multi-Layer Perceptrons (MLPs) and supports queries of density and radiance given 3D coordinates and view directions, which are used to render photo-realistic images from any view points. Moreover, during training, NeRF only requires RGB images with camera poses, which directly links 3D to 2D. The simple but ingenious architecture with its continuous representation has quickly started to challenge traditional representations using explicit discrete structures, such as RGB-D images or point clouds. As a result, NeRF is poised to tackle more challenging tasks in 3D vision. 

One important downstream task that can benefit from NeRF representations is 3D object segmentation, which is fundamental in 3D vision and widely used in many applications. Researchers have investigated applying NeRF to object segmentation. Semantic-NeRF~\cite{Zhi:etal:ICCV2021}, targeting semantic segmentation, is one of the first works in this direction. DFF~\cite{kobayashi2022distilledfeaturefields} distills the knowledge of pre-trained features such as DINO~\cite{caron2021emerging} into a 3D feature field for unsupervised object decomposition. Supervised approaches, such as \cite{Siddiqui_2023_CVPR}, utilize Mask2Former \cite{cheng2021mask2former} to obtain initial 2D masks and lifts them to 3D with a panoptic radiance field. Although these methods demonstrate impressive results, their performance is constrained by the pre-trained models used to produce features. 

Recently, large vision models such as Segment Anything Model (SAM)~\cite{kirillov2023segment} have emerged. These models with strong zero-shot generalization performance can be adopted as the backbone component for many downstream tasks. SAM presents a new paradigm for segmentation tasks. It can accept a wide variety of prompts as input, and produce segmentation masks of different semantic levels as output. The versatility and generalizability of SAM suggest a new way to perform promptable object segmentation in NeRF. While there exist some investigations~\cite{cen2023segment,chen2023san,isrfgoel2023} into this area, the mask quality in novel views is still unsatisfactory. 

In view of this, we propose a new general framework to achieve prompt-based 3D segmentation in NeRF. Our framework, termed Segment Anything for NeRF in High Quality, or SANeRF-HQ, leverages existing 2D foundation models such as Segment Anything to allow various prompts as input, and produces 3D segmentations with high accuracy and multi-view consistency. 
The major contributions of our paper are:
\begin{itemize} 
  \item We propose SANeRF-HQ, one of the first attempts at producing high-quality 3D object segmentation in NeRF in terms of more accurate segmentation boundaries and better multi-view consistency. 
  \item We validate our method by evaluating quantitatively on a challenging dataset with high-quality ground-truths.
  \item We present a general framework to embed foundation 2D image models into NeRFs and extend it to different 3D segmentation tasks in NeRFs. 
\end{itemize}

Comparing with~\cite{isrfgoel2023} and~\cite{cen2023segment}, SANeRF-HQ can produce more accurate segmentation results and is more flexible to a variety of segmentation tasks. SANeRF-HQ inherits the zero-shot performance from SAM, instead of being bounded by pre-trained models with limited generalizability ~\cite{instancenerf,bhalgat2023contrastive,Siddiqui_2023_CVPR}. Moreover, SANeRF-HQ is a general framework, which has the hidden capability to automatically segment a given 3D scene like SAM~\cite{kirillov2023segment} in 2D segmentation. Also, SANeRF-HQ can be potentially extended to 4D dynamic NeRFs, where temporal consistency can be  handled in a similar way as our multi-view consistency. Our preliminary results demonstrate promising prospect for the extension to dynamic scenes.

\section{Related Work}
\subsection{Object Segmentation}
2D image segmentation including semantic segmentation~\cite{zheng2021rethinking, guo2022segnext, chen2018encoder, zhao2017pyramid}, instance segmentation~\cite{tian2020conditional, wang2020solo, he2017mask, yolact-plus-tpami2020, yolact-iccv2019, wang2020solov2}, and panoptic segmentation~\cite{panopticsegformer, jain2023oneformer, li2019attention, kirillov2019panoptic, xiong2019upsnet, de2018panoptic} is a thoroughly studied area. With the emergence of Transformer-based models~\cite{carion2020end}, Vision Transformer (ViT)~\cite{dosovitskiy2020image, liu2021swin, yuan2021tokens, yue2021vision} has become increasingly popular as a backbone structure, and pre-trained features from large models such as MAE~\cite{he2022masked, tong2022videomae, wei2022masked}  have demonstrated great power in segmentation tasks. Meanwhile, significant research effort~\cite{cheng2021mask2former, zhang2021knet, jain2023oneformer} has focused on universal models to accomplish multiple segmentation tasks under different training configurations. However, the performance of these models on open-world images is not satisfactory, as they cannot go beyond the limits prescribed by the underlying training datasets.

In response, the latest research has been focusing on open-world segmentation, aiming to generalize segmentation models to unseen data. The recent advancement in visual foundation models have attracted great attention. DINOv2~\cite{oquab2023dinov2} leverages self-supervised distillation and produces visual features across domains without any fine-tuning. SAM~\cite{kirillov2023segment}, a more significant breakthrough, shows promising results on promptable segmentation. Given diverse and plentiful training data, the prompt-based architecture can enable zero-shot generalization, which extends relevant tasks to wider data categories. Recent works~\cite{rajivc2023segment, cheng2023tracking, zou2023segment} have already adopted SAM in many downstream tasks, enhancing the capability of different segmentation models.  

In spite of the great success of 2D segmentation, 3D segmentation is relatively underexplored. Traditional methods are usually based on RGB-D images~\cite{hou2019sis} or point clouds~\cite{vu2022softgroup, ngo2023isbnet}. However, they require explicit depth or 3D representations as input, and the generalizability is highly restricted by the scarcity of the dataset and the expensive computational cost. Therefore, exploiting the vast 2D image datasets and performing 3D segmentation directly from 2D multi-view images warrants attention from the community.

\vspace{-0.05in}
\subsection{Segmentation in Neural Radiance Field}
\label{sec:seg_in_nerf}
NeRF~\cite{mildenhall2020nerf, mueller2022instant, Chen2022ECCV, barron2022mipnerf360, barron2021mipnerf, barron2023zipnerf} has become the state-of-the-art for novel view synthesis. Beyond that, requiring only multi-view images with camera parameters during training, NeRF can also be regarded as an implicit 3D representation since it captures the 3D structural details of the scenes. Due to its capability of linking 2D images to 3D volumes, NeRF shows potential impact in various 3D visual tasks, and numerous research works have been focusing on 3D object segmentation and scene decomposition in NeRF.

\begin{figure*}[t]
\centering
    \includegraphics[width=1\linewidth]{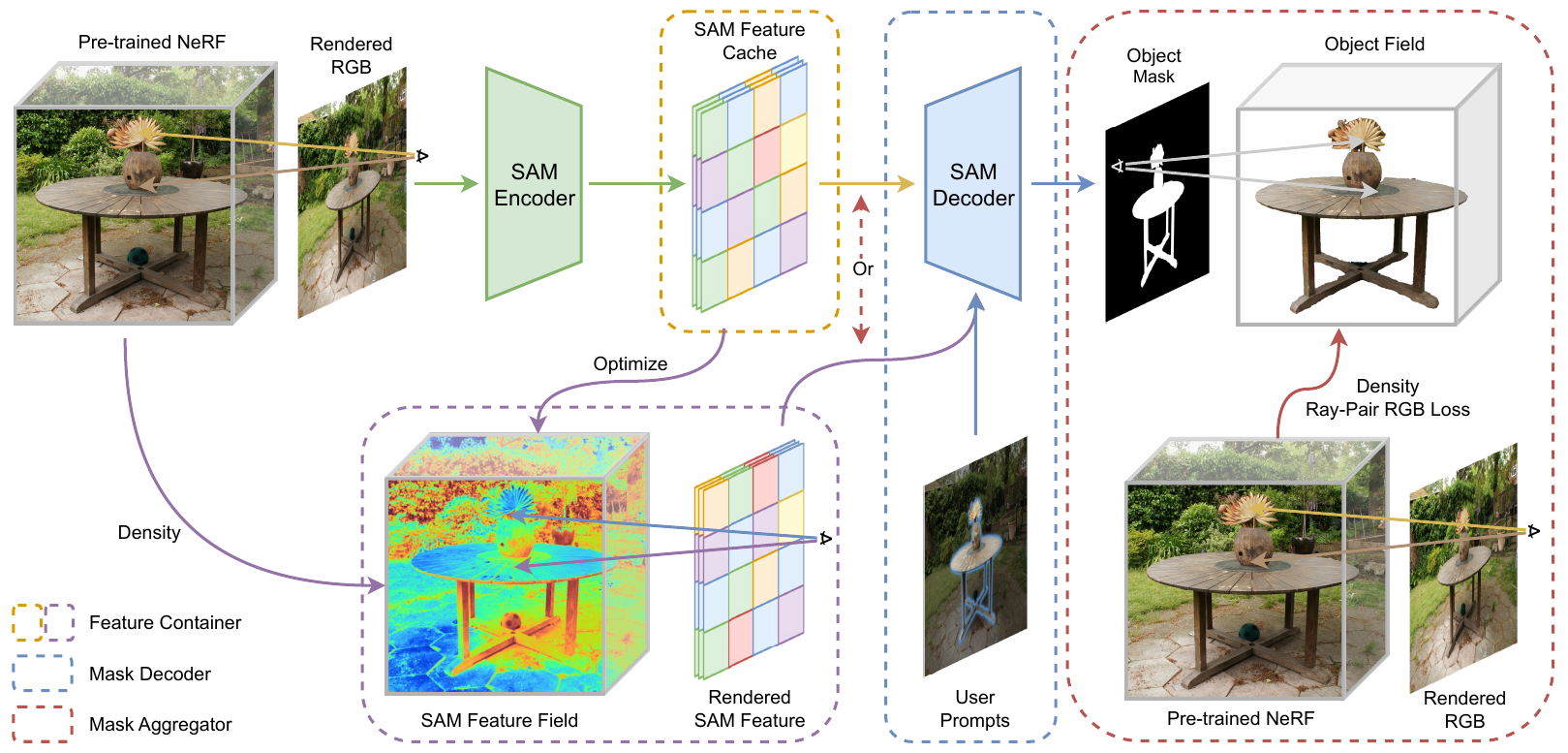}
    \vspace{-0.2in} 
    \caption{\textbf{SANeRF-HQ Pipeline.}  Our method is composed of three parts: a feature container (feature cache or feature field), a mask decoder, and a mask aggregator (object field). It first renders a set of images using a pre-trained NeRF and encodes their SAM features, which are cached or used to optimize a feature field. SAM decoder takes the feature maps from the cache or the feature field, and generates 2D masks based on user prompts. The aggregator fuses 2D masks from different views to produce an object field.\vspace{-0.15in}
    }
    \label{fig:main}
\end{figure*}

~\cite{Zhi:etal:ICCV2021, gnesf} extend NeRF with an additional branch encoding 3D semantic labels for semantic-level segmentation. Other research investigates 3D instance segmentation and makes great efforts in solving inconsistency across views. For example, \cite{instancenerf} adopts an 3D object detection method~\cite{nerfrpn} to resolve 2D mask correspondences. \cite{Siddiqui_2023_CVPR} utilizes linear assignment, while~\cite{bhalgat2023contrastive} employs a contrastive loss to optimize 3D instance embeddings. There are also unsupervised methods~\cite{stelzner2021decomposing, yu2022unsupervised, Niemeyer2020GIRAFFE}, which can separate foreground objects from background. However, all aforementioned approaches are mostly limited to simple scenes and often struggle to generalize to complex open-world problems. To fully utilize 2D features from pre-trained models, some research~\cite{lerf2023, isrfgoel2023, kobayashi2022distilledfeaturefields} introduces an extra feature field to the vanilla NeRF model, which can fuse 2D features from pre-trained models into 3D. For instance, LERF~\cite{lerf2023} splits images into patches of different sizes to obtain multi-scale CLIP~\cite{clip} feature maps that can supervise the neural field training. ISRF~\cite{isrfgoel2023} uses DINO~\cite{caron2021emerging} features and K-Means clustering to separate user-selected regions from background. However, lacking a powerful decoder, the segmentation based on these features is not accurate and sharp along the boundaries. 

With the proposal of SAM, SA3D~\cite{cen2023segment} employs SAM on the NeRF-rendered images and achieves 3D segmentation from a single-view 2D mask by self-prompting. Nonetheless this pipeline relies on the first-view mask and is susceptible to the ambiguity inherent in SAM in delineating intricate structures during its self-prompting. SAN~\cite{chen2023san} chooses to distill the SAM encoder with a neural field to render SAM feature m
aps from novel views, which are supplied to the SAM decoder to produce the segmentation. However, it can produce inconsistent masks in different views, and distilling low-resolution SAM feature maps results in aliasing in the output masks, in the form of jagged mask edges. Contrary to these approaches, our proposed method directly aggregates masks for neural field training, which naturally addresses the aliasing and consistency issues.

\section{Method}
Given a pre-trained NeRF, our method aims to segment any target object in 3D, conditional on the manual prompts and/or other user supplied inputs. The SANeRF-HQ pipeline, shown in Figure~\ref{fig:main}, consists of three major components: a feature container, a mask decoder, and a mask aggregator. The feature container encodes the SAM features of images. The mask decoder propagates the user-supplied prompts between different views and generates intermediate mask outputs using the SAM features from the container. Finally, the mask aggregator integrates the resultant 2D masks into 3D space and utilizes the color and density fields from NeRF models to achieve high-quality 3D segmentation.

\subsection{Feature Container}
\label{sec:container}
The first step of utilizing SAM is to encode the images into 2D features using the SAM encoder. These features can be used repeatedly when predicting and propagating masks, thus can be pre-computed or distilled for a scene and reused for different prompts. 

We consider two methods for the feature container. The first method is to compute and cache the features of multiple views. This allows us to reuse ground-truth SAM features for different user prompts and generate accurate 2D masks when decoding. However, the cache size is constrained by the memory available. This method also requires extra time to run the encoder if users choose to supply prompts on any of the uncached novel views.

Another method is to distill the SAM features using a neural field, which is similarly done in SAN~\cite{chen2023san} and in~\cite{isrfgoel2023, lerf2023}, where SAM, DINO, or CLIP features are lifted into 3D. Instead of radiance or density, 3D SAM embeddings are encoded in a neural field and the same volumetric rendering equation is applied to render 2D feature maps.
Specifically, vanilla NeRF~\cite{mildenhall2020nerf} is formulated as $f(\mathbf{x},\mathbf{d}; \Theta_N) = (\sigma, \mathbf{c})$, where $\mathbf{x} = (x,y,z)$ is the position of the point, $\mathbf{d} = (\theta, \phi)$ is the view direction, and $\Theta_N$ is the set of parameters of the color and density field. The RGB color at each pixel is estimated through a ray casting process:
\begin{equation}
\begin{array}{l}
\hat{\mathbf{C}}(\mathbf{r})=\sum_{k=1}^K \hat{T}\left(t_k\right) \alpha\left(\sigma\left(t_k\right) \delta_k\right) \mathbf{c}\left(t_k\right),\\
\hat{T}\left(t_k\right)=\exp \left(-\sum_{a=1}^{k-1} \sigma\left(t_a\right) \delta_a\right),  \\ 
\alpha(x)=1-\exp (-x), \\
\delta_k=t_{k+1}-t_k,
\end{array}
\end{equation}
where $\mathbf{r}(t)=\mathbf{o}+t\mathbf{d}$ is the ray emitted from the camera center passing through that pixel, and $\sigma\left(t_k\right)$ and $\mathbf{c}\left(t_k\right)$ are the volume density and color at the point $\mathbf{o}+t_k\mathbf{d}$ along the ray. 

To encode SAM features, the SAM embedding at $(\mathbf{x}, \mathbf{d})$ is defined as $\mathbf{f} = f(\mathbf{x},\mathbf{d}; \Theta_f)$, where $\Theta_f$ is the set of parameters of the feature field. The feature $\hat{\mathbf{F}}$ integrated over ray $\mathbf{r}$ is given as:
\begin{equation}
    \hat{\mathbf{F}}(\mathbf{r})=\sum_{k=1}^K \hat{T}\left(t_k\right)\alpha\left(\sigma\left(t_k\right) \delta_k\right) \mathbf{f}\left(t_k\right).\\
\end{equation}
And the feature field is optimized with the MSE loss $\mathcal{L}_f$:
\begin{equation}
    \mathcal{L}_f=\sum_{\mathbf{r} \in \mathcal{R}(\Phi)}\left\|\hat{\mathbf{F}}(\mathbf{r})-\mathbf{F}(\mathbf{r})\right\|_2^2,
\end{equation}
where $\mathcal{R}(\Phi)$ is the set of rays from the feature map $\Phi$ and $\mathbf{F}(\mathbf{r})$ is the ground-truth feature value of the ray $\mathbf{r}$.

The feature field enables efficient feature map rendering from any viewpoint , as aggregating features in the neural field is typically faster than running the original SAM encoder. However, the feature maps produced by the SAN encoder have a relatively low resolution, which can cause severe aliasing in the rendered feature maps. While this can be alleviated by augmenting the input camera views and sampling more rays during distillation, the rendered features still deteriorate after distillation. The rendered SAN features usually fail to retain accurate high-frequency spatial information along boundaries, which consequently leads to jagged mask boundaries after decoding.

Noting the complementary advantages and disadvantages of the caching method and feature distillation method, we conducted experiments on both in the ablation study.

\subsection{Mask Decoder}
\begin{figure}[t]
\centering
    \includegraphics[width=1\linewidth]{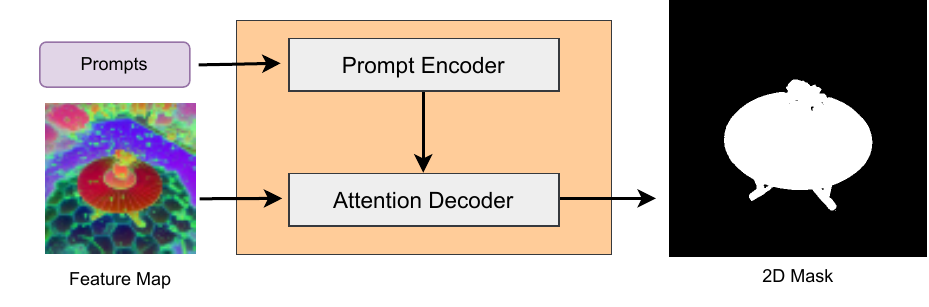}
    \vspace{-0.25in}
    \caption{\textbf{Mask Decoder Architecture.}  The decoder consists of a prompt encoder and an attention decoder. First, the prompts are fed into the prompt encoder. The attention decoder takes the encoded prompts and the feature map from the feature container, and uses attention to produce 2D masks for the given view.\vspace{-0.15in}}
    \label{fig:decoder}
\end{figure}

The mask decoder $\mathbf{Dec}$ takes as input the feature map from the feature container and generates 2D masks based on the input prompts (e.g., 2D or 3D points, texts). Figure~\ref{fig:decoder} illustrates the architecture of the decoder, which is similar to the SAM decoder. The 2D mask decoding can be formulated as 
\begin{equation}
    \mathbf{M} = \mathbf{Dec}(\hat{\Phi}, \mathit{prompts}),
\end{equation}
where $\hat{\Phi}$ is the feature map. NeRF can estimate depth with
\begin{equation} \label{eq:depth}
    \hat{\mathbf{D}}(\mathbf{r})=\sum_{k=1}^K \hat{T}\left(t_k\right) \alpha\left(\sigma\left(t_k\right) \delta_k\right) t_k, 
\end{equation}
so 3D points can be easily obtained by projecting 2D prompts from users back to 3D with camera poses. Given a 2D point $(w,h)$, its depth $d(p)$, the camera intrinsic matrix $\mathbf{K}$, and extrinsic matrix $\mathbf{P}$, the corresponding 3D point $\mathbf{p}=(x,y,z)^T$ in the world space is
\begin{equation}
    \mathbf{p} = \mathbf{P}^{-1}\mathbf{K}^{-1}\begin{pmatrix} w \cdot d(p) \\ h \cdot d(p) \\ d(p) \end{pmatrix}.
\end{equation}
The equation to project 3D points in world space to 2D pixel coordinates in other camera views can be derived likewise. 

\subsection{Mask Aggregator}
The decoder cannot produce correct 2D masks if the 3D points after projection are not visible at certain viewpoints. Furthermore, despite good performance in most cases, the predicted masks may include artifacts. The innate semantic ambiguity of SAM predictions can also cause inconsistency across views. Hence, we aggregate these imperfect 2D masks in the 3D space to generate high-quality and consistent 3D masks.
\vspace{-0.07in}
\subsubsection{Object Field}
\vspace{-0.04in}
Given the decoder output from different views, the object field can fuse all 2D images and generate accurate 3D masks. The mask is represented by an $L$-dimensional object identity vector $i$. To represent the identity value, an additional branch is introduced into the pre-trained NeRF model, parameterized as $\Theta_o$. Different from the function of RGB color, which is view-dependent, the object identity function is defined by $\mathbf{i} = f(\mathbf{x}; \Theta_o)$, where the view direction vector is not included in the inputs due to the view invariance of the object masks in 3D. The equation for mask rendering is similar to that of image rendering:
\begin{equation}
    \hat{\mathbf{M}}(\mathbf{r})=\mathit{Softmax}\left(\sum_{k=1}^K \hat{T}\left(t_k\right)\alpha\left(\sigma\left(t_k\right) \delta_k\right) \mathbf{i}\left(t_k\right)\right),\\
\end{equation}
where $\sigma(t_k)$ is inherited from the pre-trained NeRF model. The volume density $\sigma$, which interprets the 3D geometry in NeRF, makes the object field aware of the structural information. 
The object field is trained with the cross-entropy loss $\mathcal{L}_o$:
\begin{equation}\label{eq:CEL}
    \mathcal{L}_o(\mathcal{R})= -\frac{1}{|\mathcal{R}|L}\sum_{\mathbf{r} \in \mathcal{R}}\sum_{l=1}^L m^l(\mathbf{r}) \log \hat{m}^l(\mathbf{r}),
\end{equation}
where $\mathcal{R}$ is a set of rays, and $m^l(\mathbf{r})$ and $\hat{m}^l(\mathbf{r})$ are the $l$-th entry of the ground-truth mask $\mathbf{M}(\mathbf{r})$ and the predicted mask $\hat{\mathbf{M}}(\mathbf{r})$, respectively.

\vspace{-0.1in}
\subsubsection{Ray-Pair RGB Loss}
\vspace{-0.04in}
\label{sec:rgb loss}
Segmentation errors in both 3D and 2D are more likely to occur at object boundaries. One observation is that humans usually distinguish object boundaries by the color and texture difference on the two sides. Here we introduce the Ray-Pair RGB loss, aiming to incorporate color and spatial information to improve the segmentation quality.

Given a batch of rays $\mathcal{R}$, we sample a subset of rays $\mathcal{K}$ from $\mathcal{R}$ as references. For each ray $ \mathbf{r}_k \in \mathcal{K}$, we calculate the RGB similarity between $\mathbf{r}_k$ and other rays $\mathbf{r} \in \mathcal{R}$, denoted by $g(\mathbf{c}(\mathbf{r}),\mathbf{c}(\mathbf{r}_k))$, where $\mathbf{c}(\mathbf{r})$ is the rendered RGB color along $\mathbf{r}$. Next, a subset $\mathcal{S}_k$ is selected from $\mathcal{R} \setminus \mathbf{r}_k$, where for all $\mathbf{r}_s \in \mathcal{S}_k$, $g(\mathbf{c}(\mathbf{r}_s),\mathbf{c}(\mathbf{r}_k)) \geq \tau$, $\tau \in \mathbb{R}$ is a threshold. The RGB loss is defined as 
\begin{equation}
    \mathcal{L}_{RGB}(\mathcal{R})=\frac{1}{|\mathcal{K}|}\sum_{\mathbf{r}_k\in \mathcal{K}}\frac{1}{|\mathcal{S}_k|}\sum_{\mathbf{r}_s \in \mathcal{S}_k}  f(\hat{\mathbf{M}}(\mathbf{r}_k), \hat{\mathbf{M}}(\mathbf{r}_s)),
\end{equation}
where $f$ is a distance function of two probability vectors. This loss function encourages those rays with similar RGB colors to have similar object identity predictions. $\hat{\mathbf{M}}(\mathbf{r}_k)$ is detached from the compute graph, so gradients from $\mathcal{L}_{RGB}$ only flow through $\hat{\mathbf{M}}(\mathbf{r}_s)$. In our implementation, $g, f$ are defined as:
\begin{equation}
    g(\mathbf{c}_0,\mathbf{c}_1)= \|\mathbf{c}_0-\mathbf{c}_1\|_2,
\end{equation}
\begin{equation}\label{eq:distance}
    f(\mathbf{M}_0, \mathbf{M}_1)=\exp{\left(-w\frac{\mathbf{M}_0 \cdot \mathbf{M}_1}{\max (\|\mathbf{M}_0\|_2^2, \; \| \mathbf{M}_1 \|_2^2)}-\epsilon\right)}, 
\end{equation}
where $w$ and $\epsilon$ are hyperparameters.

\vspace{2mm}
\noindent\textbf{Sampling Strategy.}\quad At the beginning, only $\mathcal{L}_o$ is used to optimize the object field. Concurrently, an error map $\mathbf{E}_t$ is updated to record the difference between the rendered mask and the ground-truth mask for each training view:
\begin{equation}\label{eq:error}
    \mathbf{E}_t(\mathbf{r}) = f(\mathbf{M}(\mathbf{r}), \hat{\mathbf{M}}(\mathbf{r})),
\end{equation}
where $f$ is the function in Eq.~\ref{eq:distance}. In practice, the resolution of error maps is smaller than that of training images to reduce memory usage and increase update efficiency, so $\mathbf{E}_t(\mathbf{r})$ is approximated by $f(\mathbf{M}(\mathbf{r}'), \hat{\mathbf{M}}(\mathbf{r}'))$, where $\mathbf{r}'$ is the sample nearest to $\mathbf{r}$ in the low-resolution error map. After $k$ iterations of training, we include the Ray-Pair RGB loss $\mathcal{L}_{RGB}$ in training, which is only applied on local regions sampled according to the error maps. Specifically, a pixel $p$ from a certain viewpoint with a large error is sampled and reprojected to different viewpoints in a set $\mathcal{V}$, forming a set of pixels $\{p_v \,|\, v \in \mathcal{V}\}$ in different views. From each $p_v$, we cast a set of rays $R_{v,p}$ in the local $N\times N$ image patch around $p_v$. The entire set of rays relevant to $p$, denoted by $\mathcal{R}_p$, is defined as: 
\begin{equation}
    \mathcal{R}_p = \bigcup_{v \in \mathcal{V}} R_{v,p},
\end{equation}
on which we compute the $\mathcal{L}_{RGB}$ for $p$. This allows us to enforce the loss to rays in different views that are relevant to the same high-error region.

To maintain the global segmentation results while refining local regions, we combine Ray-Pair RGB loss with the loss function in Eq.~\ref{eq:CEL} and adopt mixed sampling: the cross entropy loss is applied to the rays sampled gloablly while the RGB loss is only applied to certain local regions. The final loss function $\mathcal{L}$ is
\begin{equation}
    \mathcal{L} = \mathcal{L}_{o}(\mathcal{R}) + \frac{1}{|\mathcal{T}|}\sum_{p \in \mathcal{T}} \mathcal{L}_{RGB}(\mathcal{R}_p),
\end{equation}
where $\mathcal{R}$ is a set of rays sampled randomly from all training views, and $\mathcal{T}$ is a set of points sampled based on error maps.
\vspace{-0.1in}

\vspace{-0.1in}
\section{Experiments}
\newcommand\bonsaiWidth{0.02cm}
\newcommand\bonsaiHeight{0.02cm}

\begin{figure*}[ht]
    \centering
    \captionsetup[subfloat]{position=top}

    \subfloat[RGB]{\includegraphics[width = 0.18\linewidth]{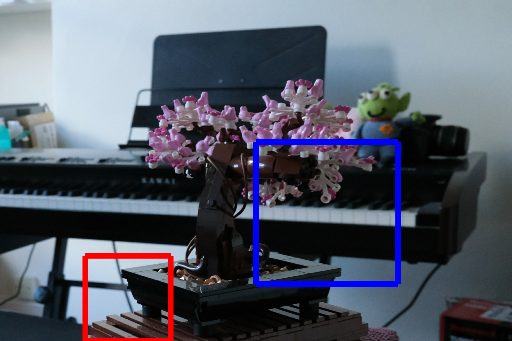}}\hspace{\bonsaiWidth}
    \subfloat[GT]{\includegraphics[width = 0.18\linewidth]{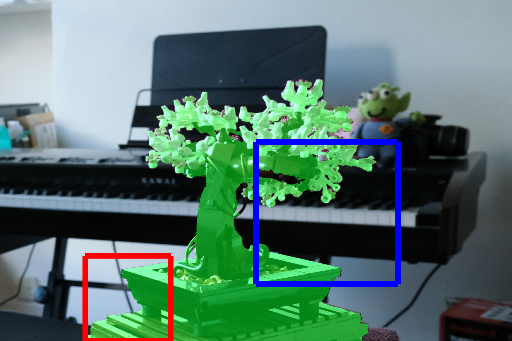}}\hspace{\bonsaiWidth}
    \subfloat[ISRF]{\includegraphics[width = 0.18\linewidth]{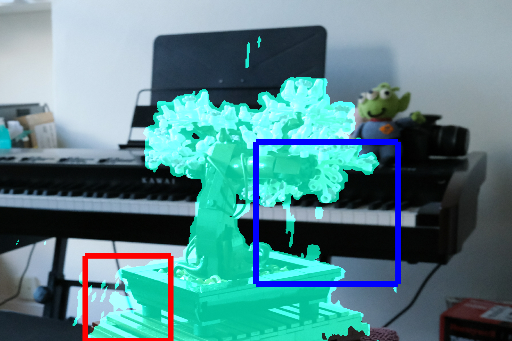}}\hspace{\bonsaiWidth}
    \subfloat[SA3D]{\includegraphics[width = 0.18\linewidth]{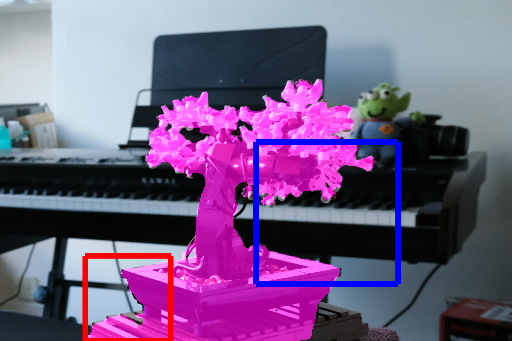}}\hspace{\bonsaiWidth}   
    \subfloat[Ours]{\includegraphics[width = 0.18\linewidth]{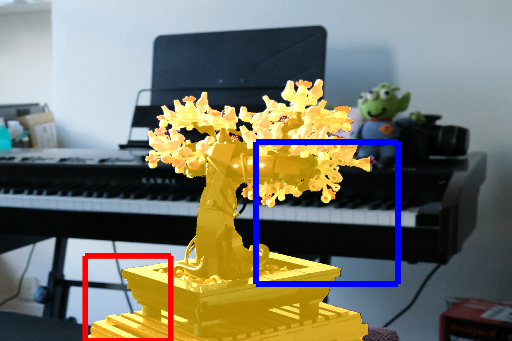}}\hspace{\bonsaiWidth}

    \vspace{\bonsaiHeight}    
    \subfloat{\includegraphics[width = 0.18\linewidth]{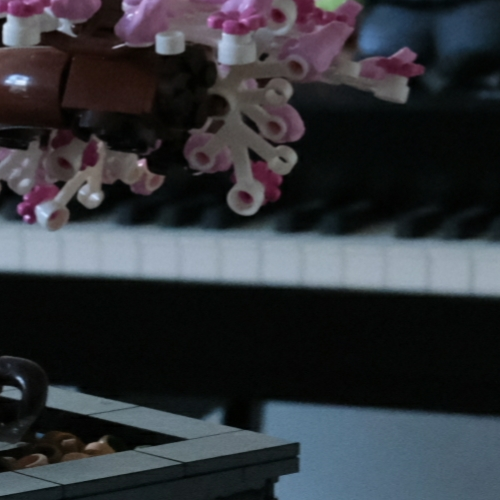}}\hspace{\bonsaiWidth}
    \subfloat{\includegraphics[width = 0.18\linewidth]{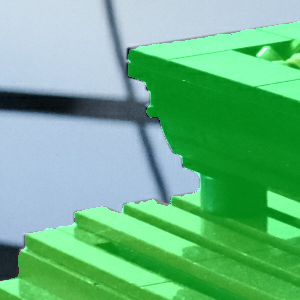}}\hspace{\bonsaiWidth}
    \subfloat{\includegraphics[width = 0.18\linewidth]{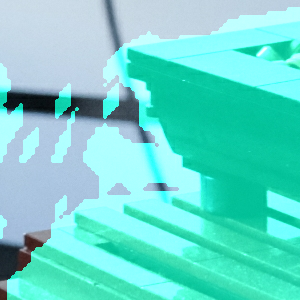}}\hspace{\bonsaiWidth}
    \subfloat{\includegraphics[width = 0.18\linewidth]{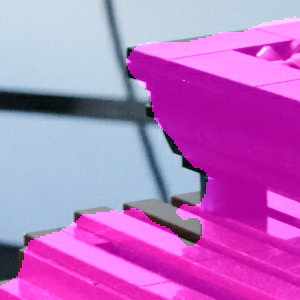}}\hspace{\bonsaiWidth}  
    \subfloat{\includegraphics[width = 0.18\linewidth]{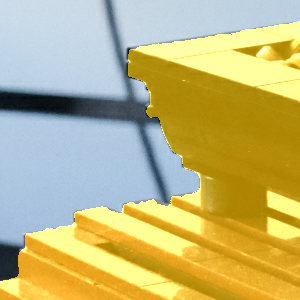}}\hspace{\bonsaiWidth}

    \vspace{\bonsaiHeight}

    \subfloat{\includegraphics[width = 0.18\linewidth]{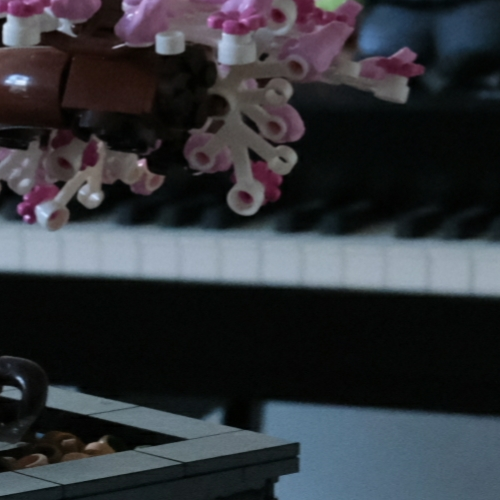}}\hspace{\bonsaiWidth}
    \subfloat{\includegraphics[width = 0.18\linewidth]{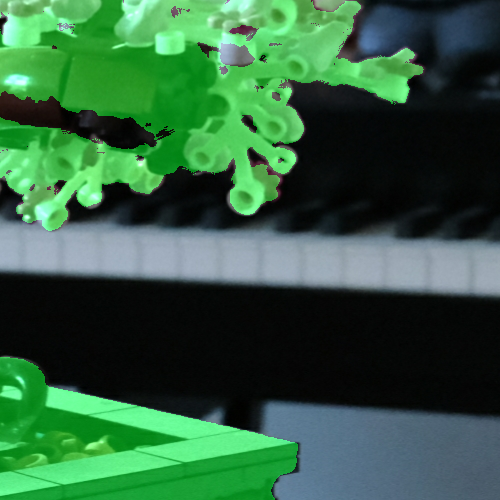}}\hspace{\bonsaiWidth}
    \subfloat{\includegraphics[width = 0.18\linewidth]{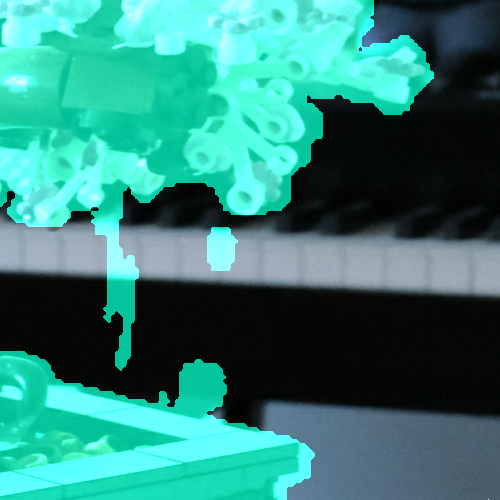}}\hspace{\bonsaiWidth}
    \subfloat{\includegraphics[width = 0.18\linewidth]{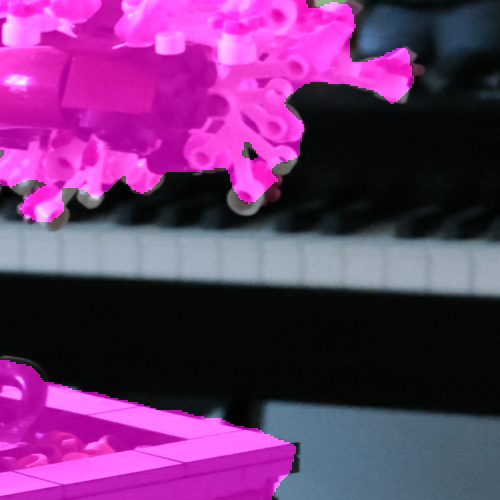}}\hspace{\bonsaiWidth}  
    \subfloat{\includegraphics[width = 0.18\linewidth]{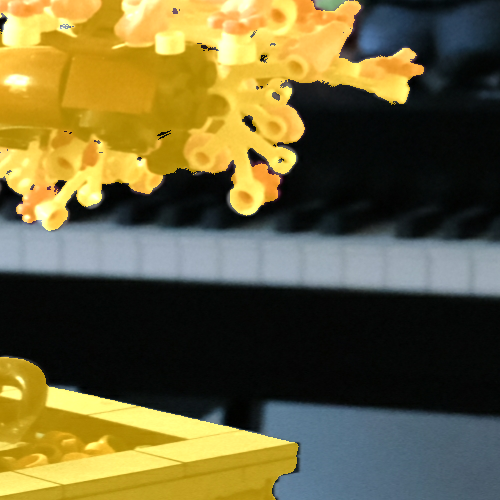}}\hspace{\bonsaiWidth}

\vspace{-0.05in}
\caption{\textbf{Comparison with SA3D and ISRF on the Bonsai.} SANeRF-HQ can produce accurate segmentation around boundaries. \vspace{-0.1in}}
    
\label{fig:comparison_3d_bonsai}
\end{figure*}

\newcommand\gardenWidth{0.02cm}
\newcommand\gardenHeight{0.02cm}
\newcommand\datasetHeight{0.03cm}
\begin{figure*}[h]
    \centering
    \captionsetup[subfloat]{position=top}

    \subfloat[RGB]{\includegraphics[width = 0.18\linewidth]{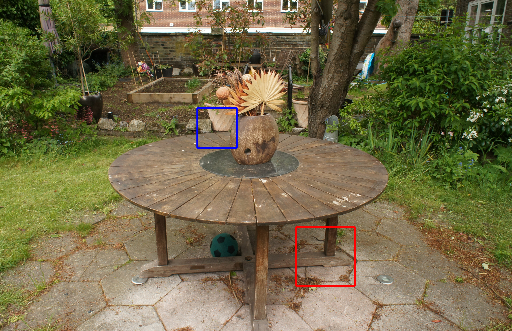}}\hspace{\gardenWidth}
    \subfloat[GT]{\includegraphics[width = 0.18\linewidth]{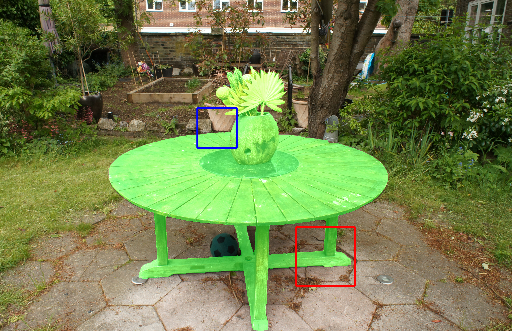}}\hspace{\gardenWidth}
    \subfloat[ISRF]{\includegraphics[width = 0.18\linewidth]{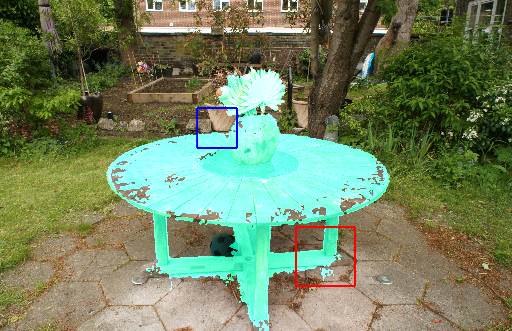}}\hspace{\gardenWidth}
    \subfloat[SA3D]{\includegraphics[width = 0.18\linewidth]{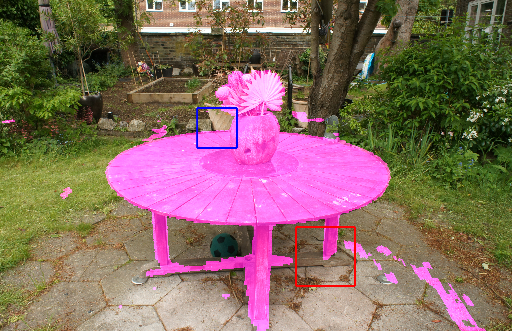}}\hspace{\gardenWidth}    
    \subfloat[Ours]{\includegraphics[width = 0.18\linewidth]{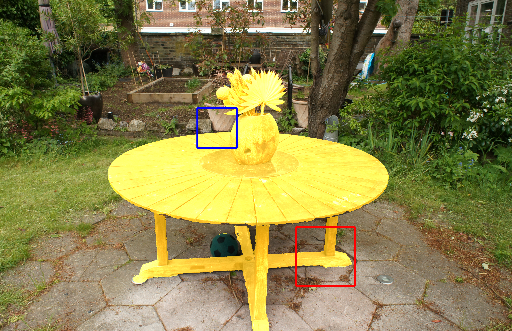}}\hspace{\gardenWidth}

    \vspace{\gardenHeight}
    \subfloat{\includegraphics[width = 0.18\linewidth]{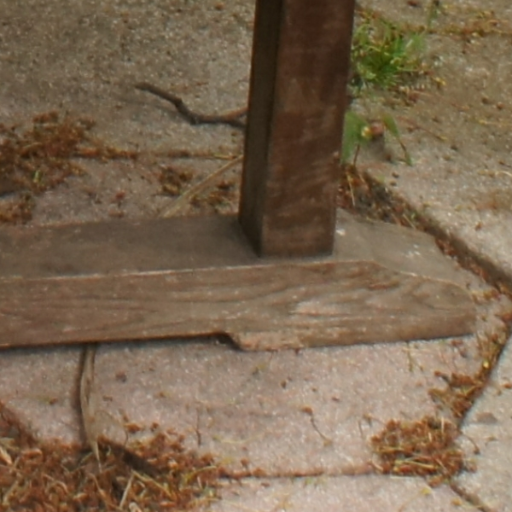}}\hspace{\gardenWidth}
    \subfloat{\includegraphics[width = 0.18\linewidth]{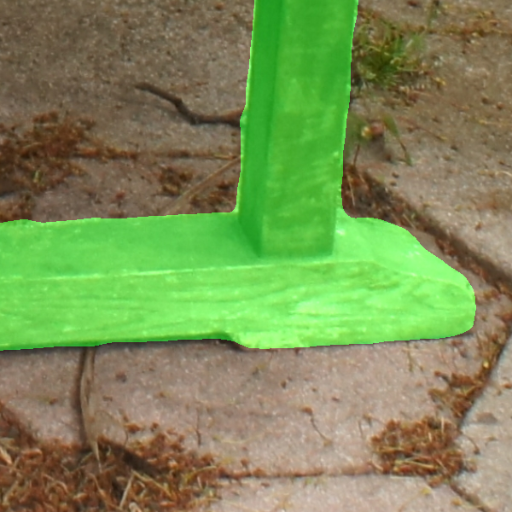}}\hspace{\gardenWidth}
    \subfloat{\includegraphics[width = 0.18\linewidth]{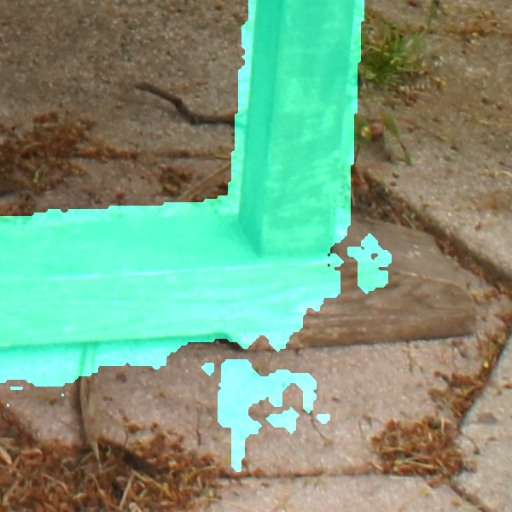}}\hspace{\gardenWidth}
    \subfloat{\includegraphics[width = 0.18\linewidth]{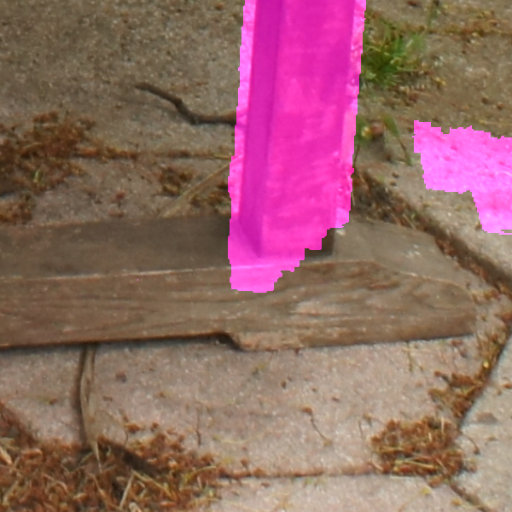}}\hspace{\gardenWidth}   
    \subfloat{\includegraphics[width = 0.18\linewidth]{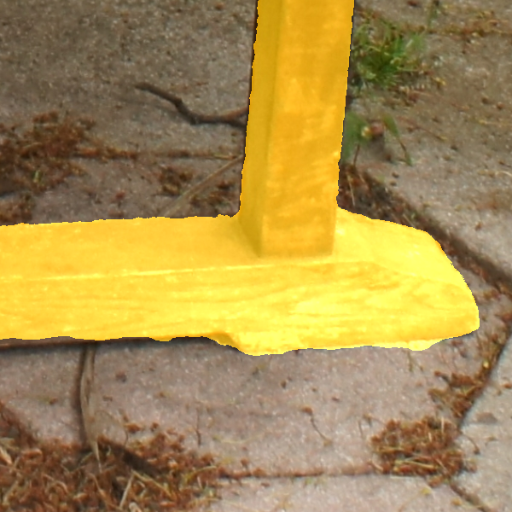}}\hspace{\gardenWidth}

    \vspace{\gardenHeight}
    \subfloat{\includegraphics[width = 0.18\linewidth]{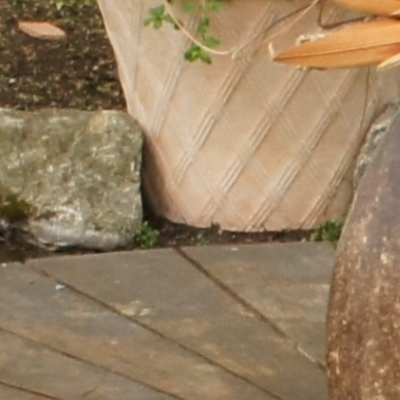}}\hspace{\gardenWidth}
    \subfloat{\includegraphics[width = 0.18\linewidth]{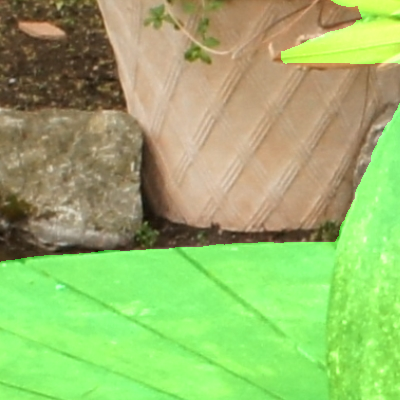}}\hspace{\gardenWidth}
    \subfloat{\includegraphics[width = 0.18\linewidth]{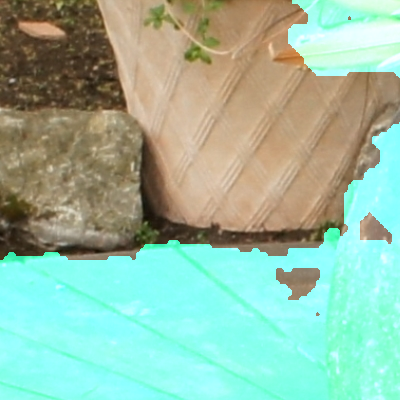}}\hspace{\gardenWidth}
    \subfloat{\includegraphics[width = 0.18\linewidth]{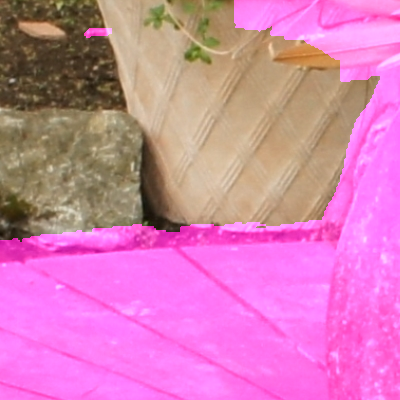}}\hspace{\gardenWidth}   
    \subfloat{\includegraphics[width = 0.18\linewidth]{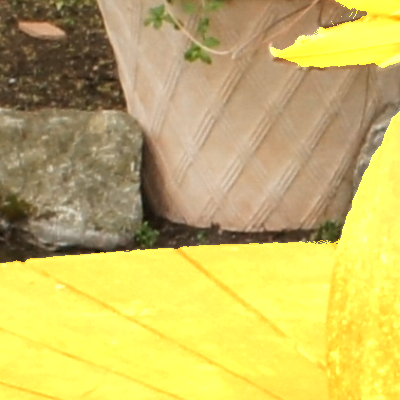}}\hspace{\gardenWidth}
    
\vspace{-0.1in}
\caption{\textbf{Comparison with SA3D and ISRF on the Garden.} SANeRF-HQ can preserve structure details of the table.\vspace{-0.2in}}

\label{fig:comparison_3d_garden}
\end{figure*}

\subsection{Datasets}\vspace{0.1in}
\label{sec:eva_set}

\newcommand\comWidth{0.02cm}
\newcommand\comHeight{0.02cm}
\newcommand\comLargeHeight{0.2cm}
\vspace{-0.1in}
\begin{figure*}[h]
    \centering
    \captionsetup[subfloat]{position=top}
    \subfloat[RGB]{\includegraphics[width = 0.18\linewidth]{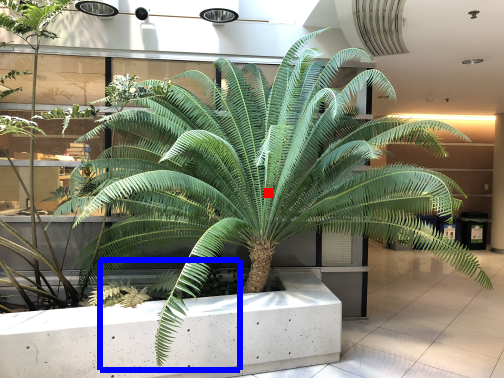}}\hspace{\comWidth}
    \subfloat[GT]{\includegraphics[width = 0.18\linewidth]{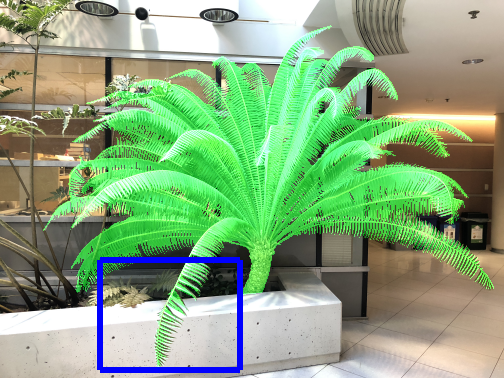}}\hspace{\comWidth}
    \subfloat[SAM]{\includegraphics[width = 0.18\linewidth]{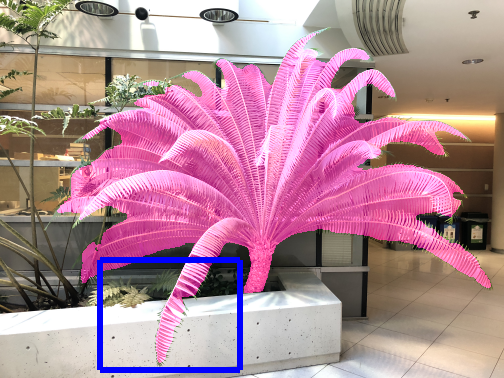}}\hspace{\comWidth}
    \subfloat[SAN]{\includegraphics[width = 0.18\linewidth]{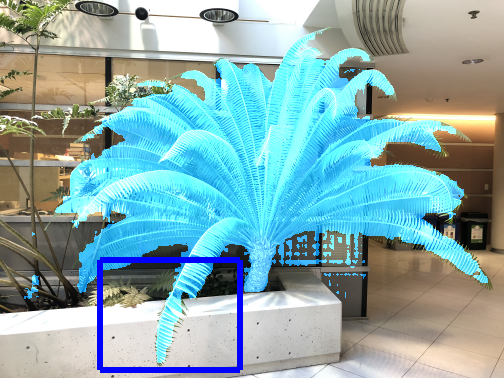}}\hspace{\comWidth}    
    \subfloat[Ours]{\includegraphics[width = 0.18\linewidth]{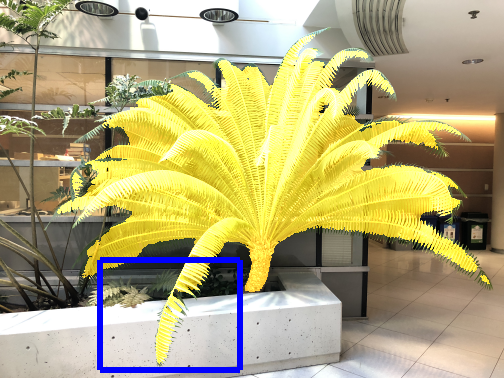}}\hspace{\comWidth}

    \vspace{\comHeight}
    
    \subfloat{\includegraphics[width = 0.18\linewidth]{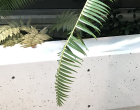}}\hspace{\comWidth}
    \subfloat{\includegraphics[width = 0.18\linewidth]{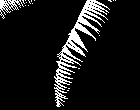}}\hspace{\comWidth}
    \subfloat{\includegraphics[width = 0.18\linewidth]{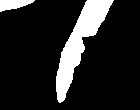}}\hspace{\comWidth}
    \subfloat{\includegraphics[width = 0.18\linewidth]{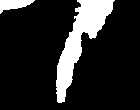}}\hspace{\comWidth}    
    \subfloat{\includegraphics[width = 0.18\linewidth]{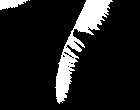}}\hspace{\comWidth}

    \vspace{\comLargeHeight}
    
    \captionsetup{position=top}
    \subfloat{\includegraphics[width = 0.18\linewidth]{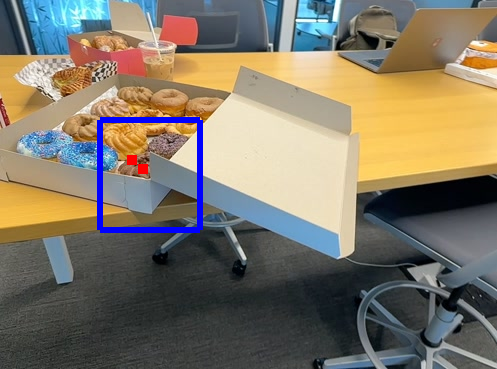}}\hspace{\comWidth}
    \subfloat{\includegraphics[width = 0.18\linewidth]{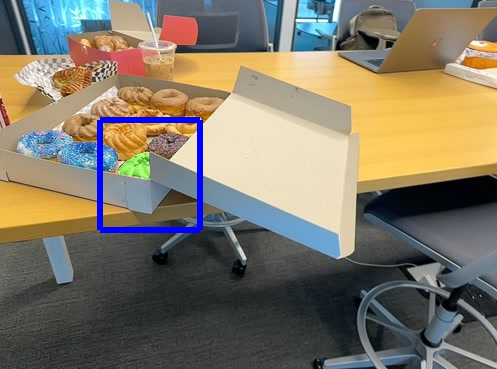}}\hspace{\comWidth}
    \subfloat{\includegraphics[width = 0.18\linewidth]{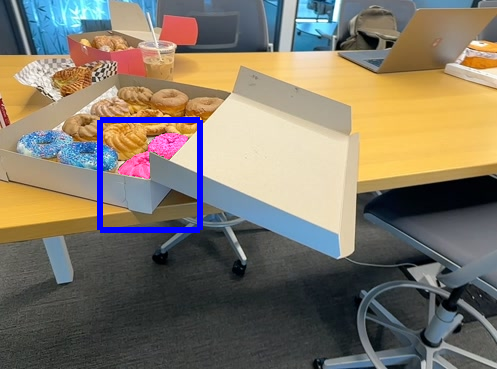}}\hspace{\comWidth}
    \subfloat{\includegraphics[width = 0.18\linewidth]{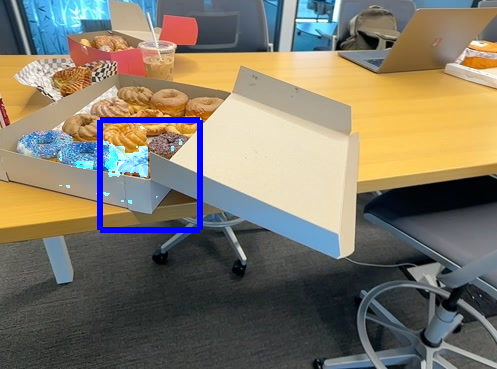}}\hspace{\comWidth}    
    \subfloat{\includegraphics[width = 0.18\linewidth]{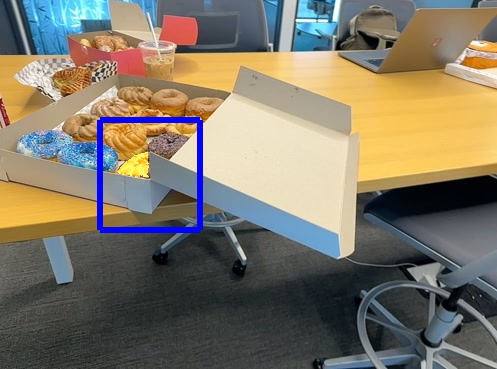}}\hspace{\comWidth}

    \vspace{\comHeight}
    
    \subfloat{\includegraphics[width = 0.18\linewidth]{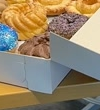}}\hspace{\comWidth}
    \subfloat{\includegraphics[width = 0.18\linewidth]{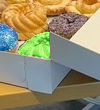}}\hspace{\comWidth}
    \subfloat{\includegraphics[width = 0.18\linewidth]{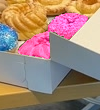}}\hspace{\comWidth}
    \subfloat{\includegraphics[width = 0.18\linewidth]{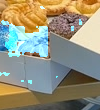}}\hspace{\comWidth}    
    \subfloat{\includegraphics[width = 0.18\linewidth]{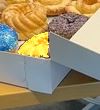}}\hspace{\comWidth}
    
\vspace{-0.05in}
\caption{\textbf{Ablation Study on the Mask Aggregator.} The red points in the RGB images represent the prompts we use in the experiments. By leveraging the 3D geometry, SANeRF-HQ can produce more accurate segmentation (the first two rows). Moreover, our method can maintain the consistency since multi-view information is fused in the object field (the last two rows). \vspace{-0.2in}}

\label{fig:comparison_2d}
\end{figure*}

We evaluate our performance on data from multiple datasets, containing synthetic and real-world scenes. We compare the masks projected on 2D images with ground-truth masks to evaluate our results quantitatively. For data without ground-truth, we manually annotate object masks. We use five datasets:
\begin{itemize}
    \item Mip-NeRF 360: a dataset widely used in NeRF research that includes synthetic and real-world examples. In our experiments, we use the data from~\cite{barron2022mipnerf360}. 
    \item LERF: a set of scenes captured in~\cite{lerf2023}, which contains complex real-world samples.
    \item LLFF: first used in~\cite{mildenhall2019llff}, the dataset contains scenes with only front views. We use the masks released in~\cite{isrfgoel2023}.
    \item 3D-FRONT: a synthetic indoor scene dataset created in~\cite{fu20213d}, further curated for NeRF training and scene understanding in Instance-NeRF~\cite{instancenerf}.
    \item Others: the rest of our evaluation set is composed of the data used in Panoptic Lifting~\cite{Siddiqui_2023_CVPR} and Contrastive Lift~\cite{bhalgat2023contrastive}. The former uses scenes from existing datasets like Hypersim~\cite{roberts:2021}, Replica~\cite{replica19arxiv} and ScanNet~\cite{dai2017scannet}, while the latter created a new dataset called Messy Rooms. 
\end{itemize}\vspace{-0.0in}
For each group above, we select scenes with representative objects and segmentation outcomes that can be clearly identified by humans. In total, 24 scenes are selected, each containing 1 to 3 object segmentations. For those without ground-truth masks, we use SAM and CascadePSP~\cite{cascadepsp} with manual annotation to create the ground-truth.

\subsection{Metrics}
We use mean Intersection over Union (mIoU) and Accuary (Acc) as the evaluation metrics. For each object, masks in novel test views are rendered and results from all test views are combined to compute the per-object IoU and Acc. The overall mIoU and Acc are averaged over all objects.

\subsection{Comparison}

We provide the comparison with some zero-shot segmentation methods mentioned in Section~\ref{sec:seg_in_nerf}. These methods leverages large vision models such as SAM~\cite{kirillov2023segment} or DINO~\cite{caron2021emerging}, and can achieve zero-shot segmentation on general scenes given user prompts. Table~\ref{tab:comparison_main} shows the quantitative comparison with four methods on 5 datasets listed in Section~\ref{sec:eva_set}. We use point prompts as they are less equivocal. For fair comparison, the same point prompts are used to get the initial masks for SA3D~\cite{cen2023segment}. As SA3D requires a single-view mask as input, we manually select a view containing the major component of the target object and pick the mask that best matches the ground-truth, instead of using the predicted scores from SAM. To our best extent, we ensure that SA3D is provided with good initialization. ISRF~\cite{isrfgoel2023} uses strokes as prompts, hence we manually connect multiple point prompts to create strokes. Figure~\ref{fig:comparison_3d_bonsai} and ~\ref{fig:comparison_3d_garden} demonstrate the qualitative comparison. Table~\ref{tab:comparison_main} presents the quantitative results.
Our method outperforms others in the comparison quantitatively on all datasets. SA3D uses a self-prompting strategy and iteratively inverse renders the 2D masks to a voxel grid, whereas our method uses a set of global prompts and collectively optimizes the object field. Despite the use of IoU rejection, self-prompting may incorrectly include occluded regions in novel views into prompts, which can accumulate errors in predicted masks, especially in initial iterations. This sensitivity to inaccurate SAM predictions partially explains SA3D's underperformance on the LERF and 3D-FRONT dataset, as the former contains scenes with small, partially-occluded objects, while the latter has furniture that may introduce semantic ambiguity.
ISRF lifts DINO features~\cite{caron2021emerging} into a neural field, but its clustering and searching process produces less accurate mask boundaries compared to SAM.

\begin{table*}
\centering
\resizebox{0.8\linewidth}{!}
{
\begin{tabular}{ccccccccccc} \toprule \multirow{2}{*}{{\textbf{Methods}}} & \multicolumn{2}{c}{{Mip-NeRF 360}} & \multicolumn{2}{c}{{LERF}} & \multicolumn{2}{c}{{LLFF}} & \multicolumn{2}{c}{{3D-FRONT}} & \multicolumn{2}{c}{{Others}} \\ \cmidrule(lr){2-3} \cmidrule(lr){4-5} \cmidrule(lr){6-7} \cmidrule(lr){8-9} \cmidrule(lr){10-11} & \small{Acc.}$\uparrow$ & \small{mIoU.}$\uparrow$ & \small{Acc.}$\uparrow$ & \small{mIoU.}$\uparrow$ & \small{Acc.}$\uparrow$ & \small{mIoU.}$\uparrow$ & \small{Acc.}$\uparrow$ & \small{mIoU.}$\uparrow$ & \small{Acc.}$\uparrow$ & \small{mIoU.}$\uparrow$ \\

\midrule  {SA3D} & \small 99.0 & \small 88.8 & \small 96.3 & \small 69.3 & \small 98.7 & \small 90.6 & \small 97.3 & \small 78.7 & \small \textbf{99.6} & \small 88.8 \\
{ISRF} & \small 95.2 & \small 65.7 & \small 88.5 & \small 27.9 & \small 96.7 & \small 80.0 & \small 92.4 & \small 68.5 & \small 86.5 & \small 23.6 \\

\midrule {SAM} & \small 97.9 & \small 80.4 & \small 98.0 & \small 82.9 & \small 99.1 & \small 93.7 & \small 97.4 & \small 77.7 & \small 99.2 & \small 83.6 \\
{SAN} & \small 97.6 & \small 77.2  & \small 98.1 & \small 71.0 & \small 96.7 & \small 83.0 & \small 97.0 & \small 76.8 & \small 98.4 & \small 73.0 \\

\midrule {Ours} & \small\textbf{99.2} & \small\textbf{91.0} & \small\textbf{99.0} & \small\textbf{90.7} & \small\textbf{99.3} & \small\textbf{95.2} & \small\textbf{98.6} & \small\textbf{89.9} & \small\textbf{99.6} & \small\textbf{91.1} \\
 
\bottomrule \end{tabular}
}
\vspace{-0.08in}
\caption{\textbf{Quantitative Results on Different Datasets.}\vspace{-0.1in}}
\label{tab:comparison_main} 
\end{table*}

\begin{table*}
\centering
\resizebox{0.88\linewidth}{!}{%
\begin{tabular}{cccccccccccc} \toprule \multirow{2}{*}{\textbf{Backbone}} & \multirow{2}{*}{\textbf{Container}} & \multicolumn{2}{c}{{Mip-NeRF 360}} & \multicolumn{2}{c}{{LERF}} & \multicolumn{2}{c}{{LLFF}} & \multicolumn{2}{c}{{3D-FRONT}} & \multicolumn{2}{c}{{Others}} \\ \cmidrule(lr){3-4} \cmidrule(lr){5-6} \cmidrule(lr){7-8} \cmidrule(lr){9-10} \cmidrule(lr){11-12} & &  \small{Acc.}$\uparrow$ & \small{mIoU.}$\uparrow$ & \small{Acc.}$\uparrow$ & \small{mIoU.}$\uparrow$ & \small{Acc.}$\uparrow$ & \small{mIoU.}$\uparrow$ & \small{Acc.}$\uparrow$ & \small{mIoU.}$\uparrow$ & \small{Acc.}$\uparrow$ & \small{mIoU.}$\uparrow$ \\
\midrule
\multirow{2}{*}{SAM} & {Cache} &  \small{99.3} &  \small{93.6} & \small {98.9} & \small{90.3} & \small{99.5} & \small{95.8} & \small{97.6} & \small{84.8} & \small\textbf{99.6} & \small\textbf{91.1}  \\
 & {Distillation} & \small{99.2} & \small{91.0} & \small{99.0} & \small{90.7} & \small{99.3} & \small{95.2} & \small{98.6} & \small{89.9} & \small\textbf{99.6} & \small\textbf{91.1} \\
\midrule
{SAM-HQ} & {Cache} & \small\textbf{99.4} & \small\textbf{94.4} & \small\textbf{99.2} & \small\textbf{93.1} & \small\textbf{99.6} & \small\textbf{96.8} & \small\textbf{98.9} & \small\textbf{91.8} & \small{99.5} & \small{90.1}  \\
\bottomrule 
\end{tabular}%
}
\vspace{-0.12in}
\caption{\textbf{Quantitative Results on Different Backbones and Feature Containers.} \vspace{-0.22in}}
\label{tab:comparison_backbone_container} 

\end{table*}

\subsection{Ablation Studies}

\noindent\textbf{SAM Models and Feature Containers.}\quad In Section~\ref{sec:container}, two methods to store the SAM features are proposed. We investigate the performance of these two methods, as well as two pre-trained segmentation models, original SAM and HQ-SAM~\cite{sam_hq}, on our evaluation set. Table~\ref{tab:comparison_backbone_container} demonstrates the quantitative results. HQ-SAM with a cache container performs the best in most cases, whereas HQ-SAM with distillation is not included due to heavier overhead of distilling multi-level feature maps. We find no significant performance gap between feature distillation and caching over different datasets, hence the difference in computational cost may play a more important role, which we discuss in the supplementry material.

\noindent\textbf {Mask Aggregator.}\quad Based on point prompts, we perform ablation study on the mask aggregator, comparing the intermediate results from the feature container, i.e., directly decoding the SAM features from the encoder or the feature field, to the masks after aggregating. The quantitative results are in Table~\ref{tab:comparison_main}. 

Directly propagating point prompts between different views and applying SAM cannot guarantee cross-view consistency. When prompts are sparse, some 3D point prompts may be occluded at certain viewpoints and no mask can be produced. Even when a large number of prompts are provided, SAM masks may still cover different objects across views, and this naive approach fail to utilize masks from other views to collectively refine the results. The same issue also exists in SAN, which distills SAM features with another neural field and later decodes the rendered features using the decoder. In addition to the consistency issue, the rendered feature maps from SAN further suffer from aliasing, and fine spatial semantics in the SAM features along object boundaries can be lost during interpolation. This can lead to less accurate segmentation results on the object boundaries, like jagged mask edges. To ensure fairness in comparison, we introduce enough point prompts to guarantee masks can be generated from every viewpoint where the target object is visible. 

Figure~\ref{fig:comparison_2d} illustrates the qualitative comparison results. Although SANeRF-HQ uses the potentially inconsistent segmentation from SAM as input, by aggregating multi-view information and integrating 3D geometry captured by NeRF, it can produce a underlying 3D mask close to the ground-truth geometry. This guarantees consistent multi-view masks and usually comes with higher quality. 

\label{sec:rationale}
\noindent\textbf {Ray-Pair RGB Loss.}\quad Table~\ref{tab:ablation_loss} shows the quantitative results of ablations on the Ray-Pair RGB loss in Section~\ref{sec:rgb loss}, where the Ray-Pair RGB loss slightly enhances the mask quality. We refer readers to the supplementary material for qualitative comparison.
\vspace{-0.05in}
\begin{table}[ht]
\centering
\resizebox{0.95\linewidth}{!}{
\vspace{0.2in}
\begin{tabular}{lcccccc}
\hline
\textbf{Metrics} & {$\mathcal{L}_{RGB}$} & {Mip-NeRF} & {LERF} & {LLFF} & {3D-FRONT} & {Others}  \\\hline
\multirow{2}{*}{mIoU.$\uparrow$} & w/o  &  91.0 & 88.3 & 95.2 & 89.3 & 90.9\\ 
& w/ & \textbf{91.3} & \textbf{90.7} & \textbf{95.8} & \textbf{89.9} & \textbf{91.1} 
\\ \hline 
\multirow{2}{*}{Acc.$\uparrow$} & w/o  & \textbf{99.2} & 98.9 & 99.4 & 98.6 & \textbf{99.6} \\
 & w/  & \textbf{99.2} & \textbf{99.0} & \textbf{99.5} & \textbf{98.7} & \textbf{99.6} 
  \\ \hline
\end{tabular}
}
\vspace{-0.15in}
\caption{\textbf{Ablation Results of the Ray-Pair RGB Loss.} \vspace{-0.2in} }
\label{tab:ablation_loss}

\end{table}

\subsection{More Qualitative Results}
Figure~\ref{fig:qual_results} illustrates the results of our methods on other segmentation tasks. With Grounding-DINO~\cite{grounding-dino}, our method can segment objects based on text prompts. Additionally, our method can perform automatic 3D segmentation, utilizing videos rendered by NeRF, the auto-segmentation function of SAM, and incorporating~\cite{cheng2023tracking} into the mask decoder. 

More qualitative results, comparisons, preliminary results on dynamic NeRF segmentation, and discussion on limitations can be found in the supplementary material.

\newcommand\resultsWidth{-0.00cm}
\newcommand\resultsHeight{0.05cm}

\vspace{-0.1in}
\begin{figure}[h]
    \centering
    \captionsetup[subfloat]{position=top, labelformat=empty}

    \subfloat[Text: \textbf{``fossil"}]{\includegraphics[width = 0.31\linewidth]{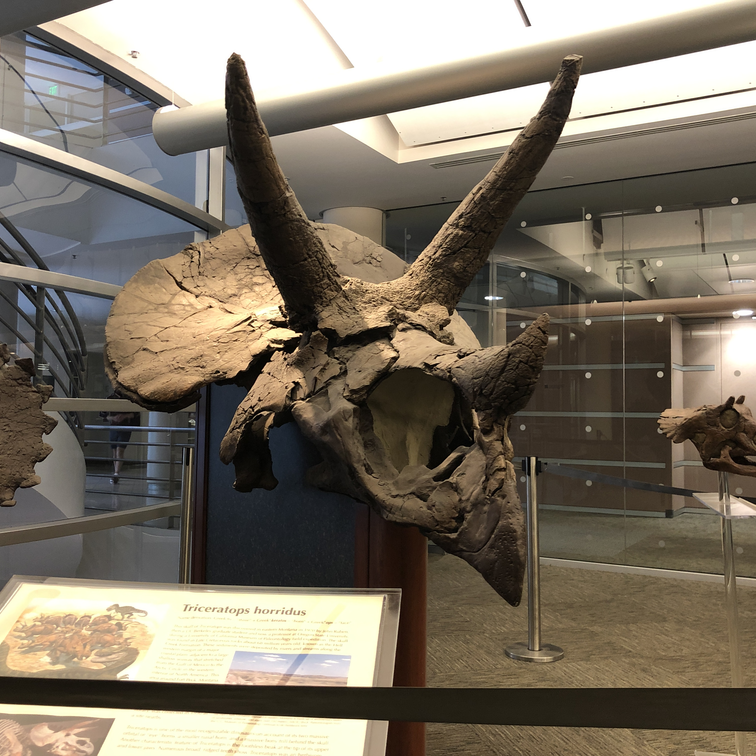}}\hspace{\resultsWidth}
    \subfloat[Text: \textbf{`` table"}]{\includegraphics[width = 0.31\linewidth]{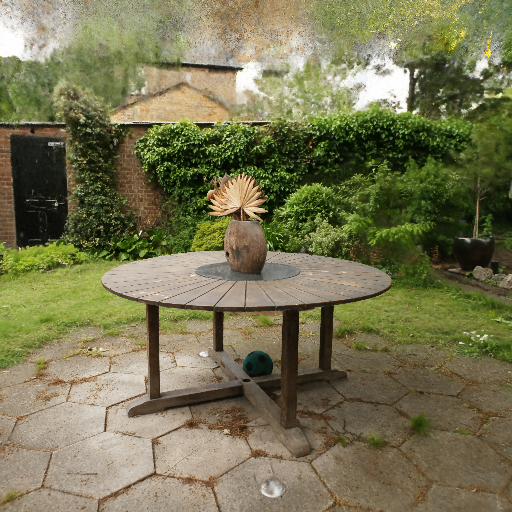}}\hspace{\resultsWidth}
    \subfloat[Auto-Seg]{\includegraphics[width = 0.31\linewidth]{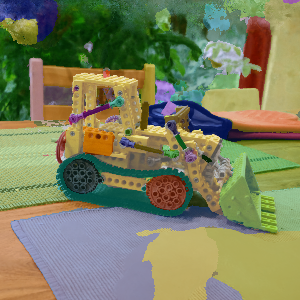}}\hspace{\resultsWidth}

    \vspace{\resultsHeight}

    \subfloat{\includegraphics[width = 0.31\linewidth]{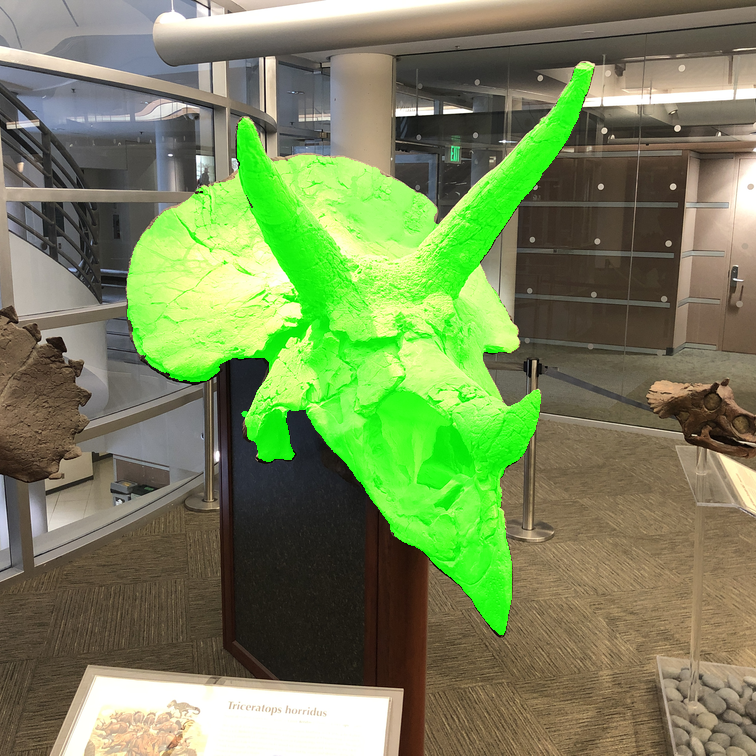}}\hspace{\resultsWidth}
    \subfloat{\includegraphics[width = 0.31\linewidth]{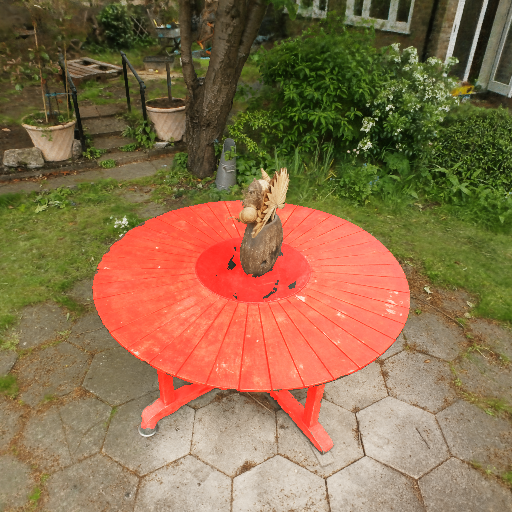}}\hspace{\resultsWidth}
    \subfloat{\includegraphics[width = 0.31\linewidth]{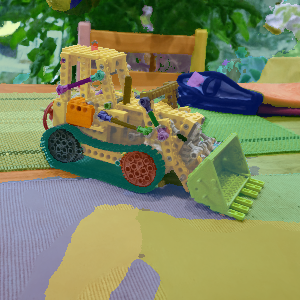}}\hspace{\resultsWidth}
    
    \vspace{-0.1in}
    \caption{ \textbf{Qualitative Results of Text Prompts and Auto Segmentation. \vspace{-0in} } 
    }
    \label{fig:qual_results}
\end{figure}
\vspace{-0.3in}

\section{Conclusion}
\vspace{-0.05in}
\begin{spacing}{1}
{In this paper, we introduced the Segment Anything for NeRF in High Quality framework. By combining the strengths of SAM for open-world object segmentation and NeRF for aggregating information from multiple viewpoints, SANeRF-HQ represents a significant advancement in high-quality 3D segmentation. Our method was quantitatively and qualitatively evaluated on various NeRF datasets, which demonstrates SANeRF-HQ's advantages over previous methods. In supplementary material, we demonstrate the potential of extending our work to object segmentation in 4D dynamic NeRFs. SANeRF-HQ holds promise for contributing significantly to the evolving landscape of 3D computer vision and segmentation techniques.}
\end{spacing}
{
    \small
    \bibliographystyle{ieeenat_fullname}
    \bibliography{main}
}

\appendix
\counterwithin{figure}{section}
\counterwithin{table}{section}
\clearpage

\section{Implementation Details}
We use torch-ngp \cite{torch-ngp} as our initial NeRF implementation. When we use 3D points as prompts in evaluation, the views containing less than $k$ visible points get filtered out automatically and will not be used to train the object field, where $k$ is a hyperparameter, depending on the total number of input points.  

For both the SAM feature field and the object field, we use a hash grid as in \cite{mueller2022instant} with 16 levels and feature dimension of 8 per level. The lowest and highest level are of resolution 16 and $2^{19}$, respectively. We use a 5-layer 256-hidden dimensional MLP with skip connections and Layer Normalization after the feature field hash grid, and a 3-layer 256 hidden dimensional MLP with skip connections after the object field hash grid. In addition to the features from their respective hash grid, both MLPs also take the features from the density field as input, where feature MLP also takes the viewing directions as input. The initial radiance and density field, the SAM feature field, and the object field are trained for 15,000, 5,000, and 600 iterations, respectively. All models are trained on an NVIDIA RTX 4090 GPU.

Ray-Pair RGB loss is included after 300 iterations of warm-up. We use error maps downsampled by 4 times compared to original training images for Ray-Pair RGB loss sampling. In each iteration, we update the error maps using the training ray batch, and for every 200 iterations, we perform a full update for all error map pixels. During sampling, we independently sample initial rays on each error map weighted by their errors, reproject them onto each view, and subsequently sample 32 additional rays in each $N \times N$ patch centered at the reprojected pixels randomly. Here we choose $N = 8$ or 16. A subset of 20 rays are then sampled from each set as references in the Ray-Pair RGB loss.

\section{Efficiency Evaluation}
To provide a more comprehensive understanding of the two methods storing SAM features mentioned in Section \ref{sec:container}, we evaluate the efficiency of the feature distillation method and the caching method. We randomly sample three scenes from the Mip-NeRF 360 dataset as reference. By default, the pre-trained NeRF renders images at 512 $\times$ 512 as input to the SAM encoder. Under a batch size of 4,096 and a maximum iteration of 5,000, it requires on average 666.0 seconds to train the feature field for a single scene, which can then render feature maps at 64 $\times$ 64 resolution from any viewpoints at 22.4 frames per second (FPS). In contrast, the caching method can encode the images to feature maps at 3.78 FPS while using extra memory to store the SAM feature maps (64 $\times$ 64, around 4.1MB each frame). Different from encoding, the decoding process is much faster, at 168.9 FPS with the pre-computed feature maps.

\section{Comparison with Instance Segmentation Methods}
We also compare our method with some instance segmentation methods. The instance segmentation methods in NeRF mentioned in our related works do not require user prompts and can automatically generate segmentation of salient objects in NeRF. These methods also leverage 2D segmentation methods for NeRF training but they mainly focus on the challenge of 3D consistency. Despite their different configurations and issues of concern, we still provide the comparison with these automatic end-to-end pipelines, showing that our prompt-based method can produce comparable results to these state-of-the-art auto-segmentation methods. Instance-NeRF~\cite{instancenerf} is a training-based methods so we only compare with it on 3D-FRONT dataset. Figure~\ref{fig:comparison_instancenerf} and Table~\ref{tab:comparison_inerf} illustrates the visual results and quantitative comparison respectively. For Panoptic Lifting~\cite{Siddiqui_2023_CVPR} and Contrastive Lift~\cite{bhalgat2023contrastive}, we also compare on the scenes they mentioned in the papers to ensure the fairness. Results are shown in Figure~\ref{fig:comparison_lift} and Table~\ref{tab:comparison_lift}.

We use the objects in our evaluation sets as targets and choose the object that has the largest IoU with the target object as the predicted results of the instance segmentation methods. Notice that we only compare with those methods on the datasets mentioned in their papers, since they do not leverage SAM to achieve zero-shot generalization.

\begin{table}[ht]
\centering
\resizebox{0.6\linewidth}{!}{
\begin{tabular}{lcc}
\hline
\textbf{Metrics} & Ours & Instance-NeRF \\
\hline \hline 
Acc.$\uparrow$ & 98.7 & 99.2 \\
mIoU.$\uparrow$ & 89.9 & 92.8 \\
\hline
\end{tabular}
}\vspace{-0.05in}
\caption{\textbf{Comparison with Instance-NeRF on 3D-FRONT}.}
\label{tab:comparison_inerf}
\end{table}
\begin{table}[ht]
\centering
\resizebox{0.70\linewidth}{!}{
\begin{tabular}{lccc}
\hline

\multirow{2}{*}{\textbf{Metrics}} & \multirow{2}{*}{Ours} & \small{Panoptic} & \small{Contrastive}  \\
 &  & \small{Lifting} & \small{Lift} \\ \hline
\hline 
Acc.$\uparrow$ & 99.6 & 94.3 & 94.1 \\
mIoU.$\uparrow$ & 91.1 & 84.5 & 81.5 \\
\hline
\end{tabular}
}\vspace{-0.05in}
\caption{\textbf{Comparison with Panoptic Lifting and Contrastive Lift.} The results are on the data mentioned in their papers.}
\label{tab:comparison_lift}
\end{table}

\newcommand\instancenerfwidth{-0.00cm}
\newcommand\instancenerfheight{0.00cm}

\vspace{-0.05in}
\begin{figure}[ht]
    \centering
    \captionsetup[subfloat]{position=top, labelformat=empty}

    \subfloat[GT Mask]{\includegraphics[width = 0.32\linewidth]{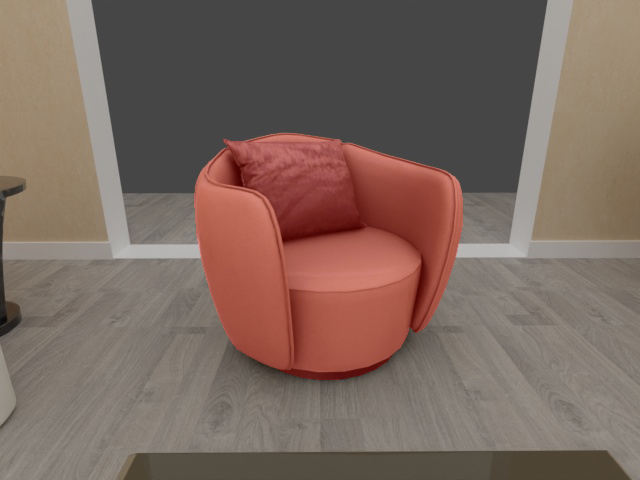}}\hspace{\instancenerfwidth}
    \subfloat[Instance-NeRF]{\includegraphics[width = 0.32\linewidth]{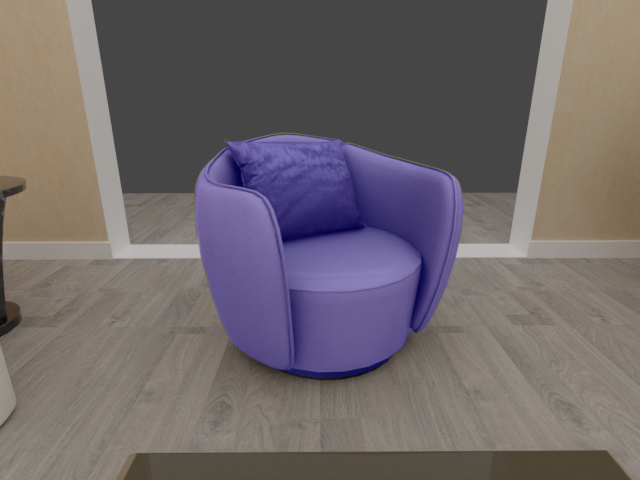}}\hspace{\instancenerfwidth}
    \subfloat[Ours]{\includegraphics[width = 0.32\linewidth]{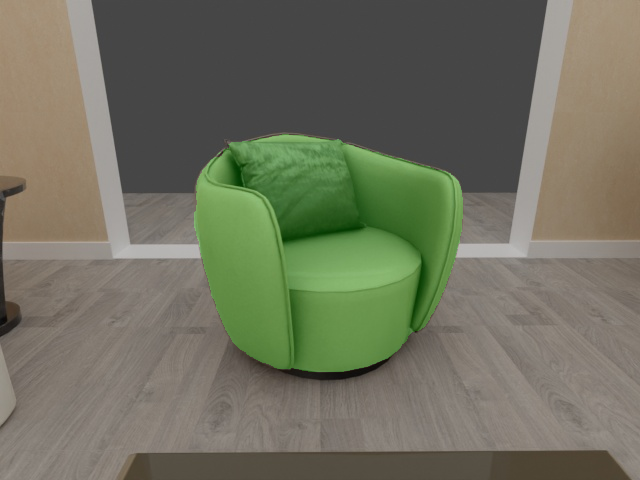}}\hspace{\instancenerfwidth}

    \vspace{\instancenerfheight}

    \subfloat{\includegraphics[width = 0.32\linewidth]{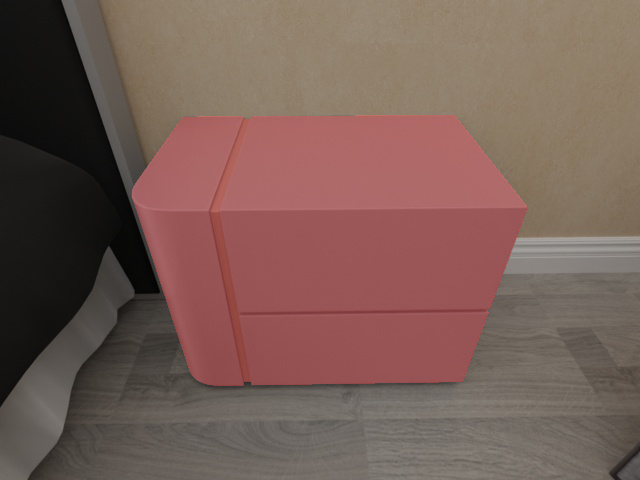}}\hspace{\instancenerfwidth}
    \subfloat{\includegraphics[width = 0.32\linewidth]{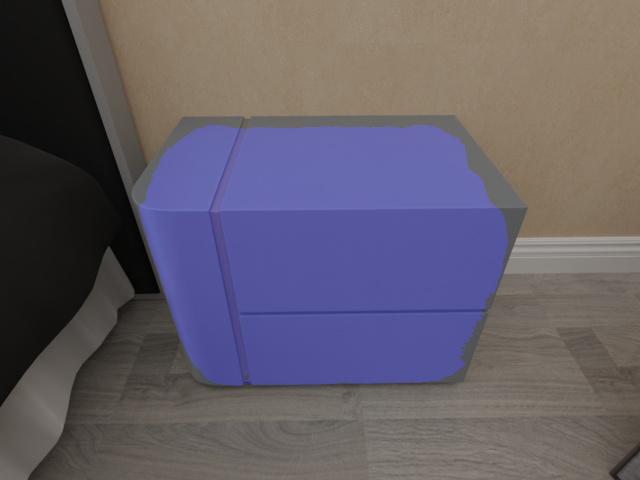}}\hspace{\instancenerfwidth}
    \subfloat{\includegraphics[width = 0.32\linewidth]{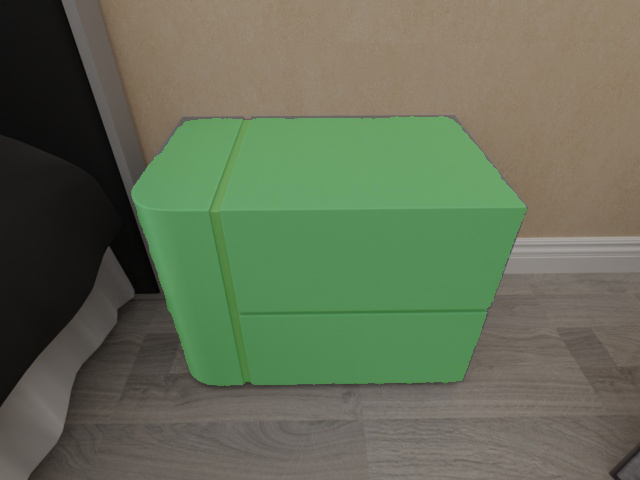}}\hspace{\instancenerfwidth}
	
    \vspace{-0.05in}
    \caption{\textbf{Qualitative Comparison with Instance-NeRF.} Zoom in for details especially along the segmentation boundaries.    
    }

    \label{fig:comparison_instancenerf}
\end{figure}
\newcommand\liftwidth{-0.00cm}
\newcommand\liftheight{0.05cm}

\vspace{-0.15in}
\begin{figure}[ht]
    \centering
    \captionsetup[subfloat]{position=top, labelformat=empty}

    \subfloat[GT Mask]{\includegraphics[width = 0.24\linewidth]{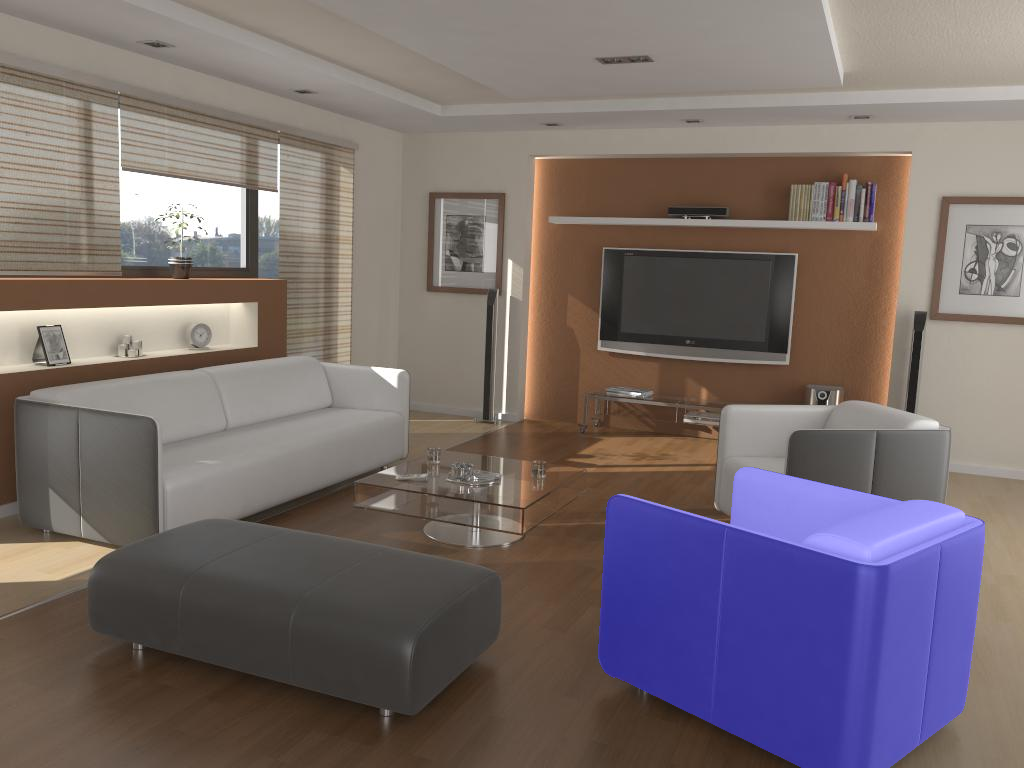}}\hspace{\liftwidth}
    \subfloat[Panoptic Lifting]{\includegraphics[width = 0.24\linewidth]{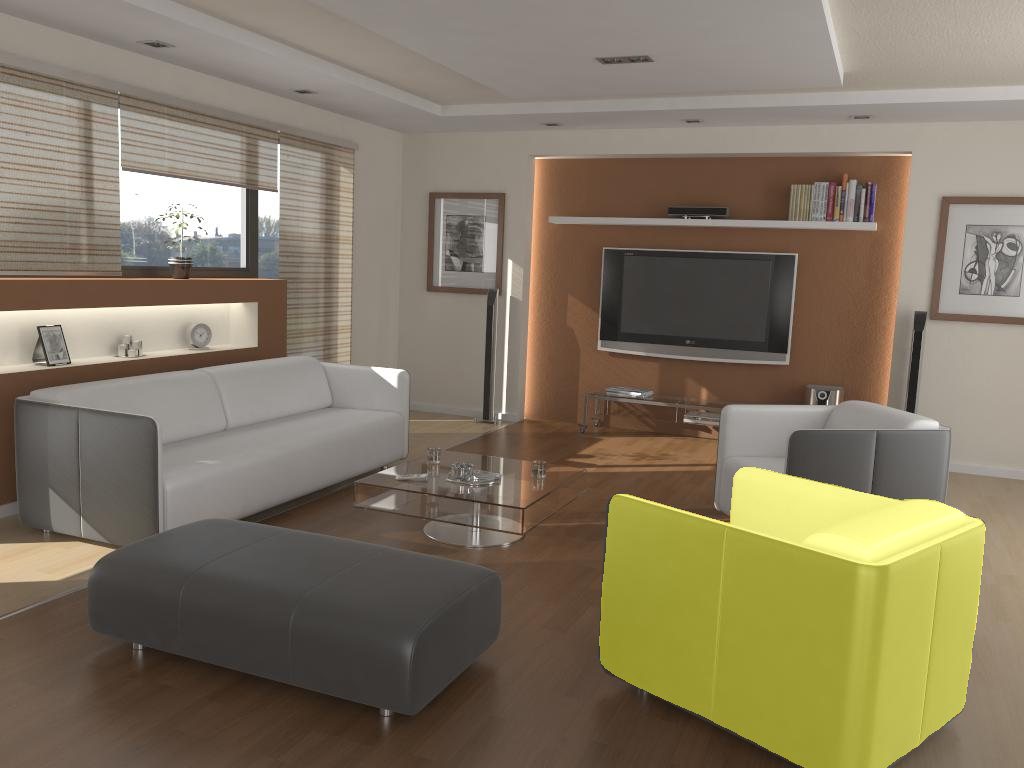}}\hspace{\liftwidth}
    \subfloat[Contrastive Lift]{\includegraphics[width = 0.24\linewidth]{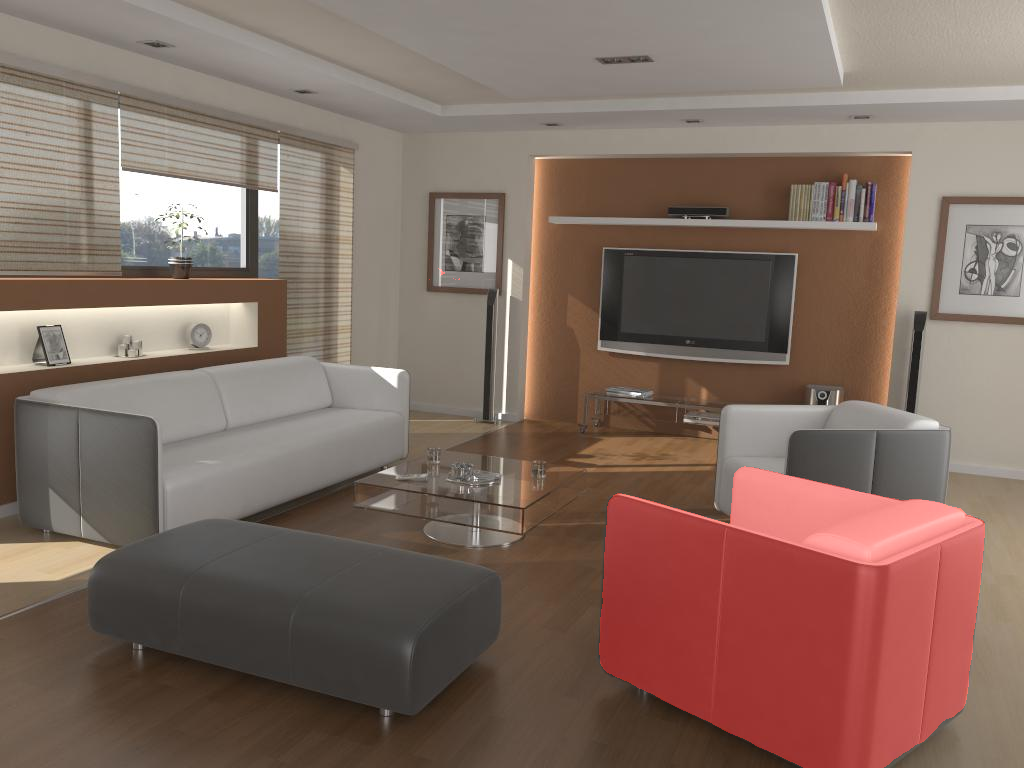}}\hspace{\liftwidth}
    \subfloat[Ours]{\includegraphics[width = 0.24\linewidth]{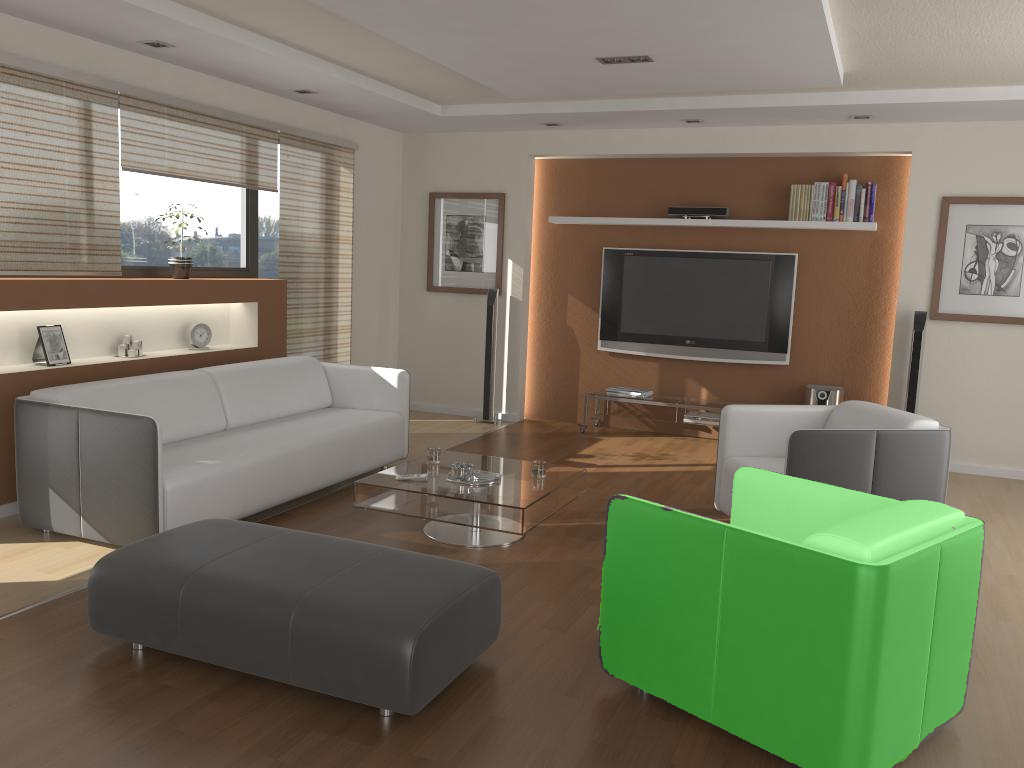}}\hspace{\liftwidth}

    \vspace{\liftheight}

    \subfloat{\includegraphics[width = 0.24\linewidth]{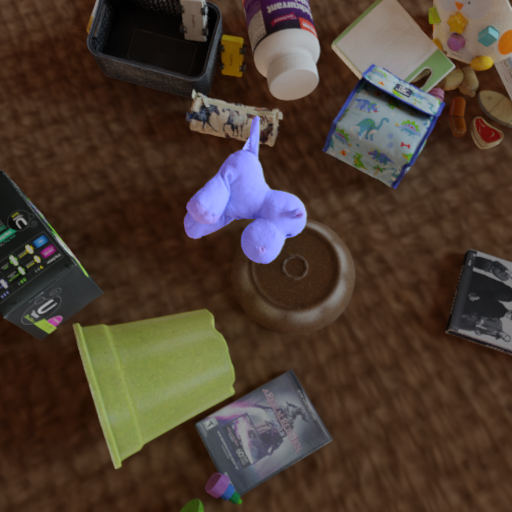}}\hspace{\liftwidth}
    \subfloat{\includegraphics[width = 0.24\linewidth]{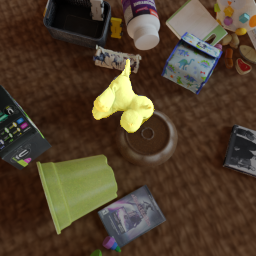}}\hspace{\liftwidth}
    \subfloat{\includegraphics[width = 0.24\linewidth]{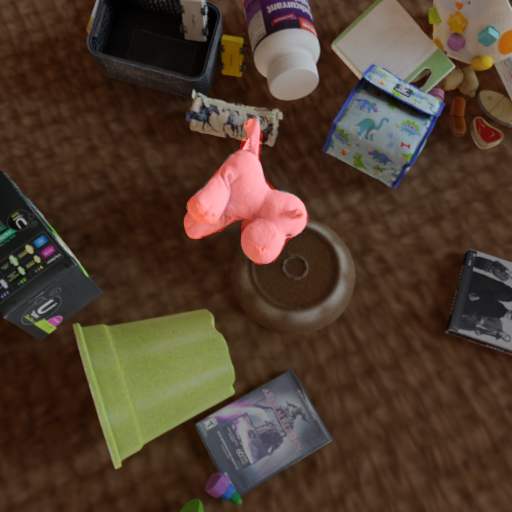}}\hspace{\liftwidth}
    \subfloat{\includegraphics[width = 0.24\linewidth]{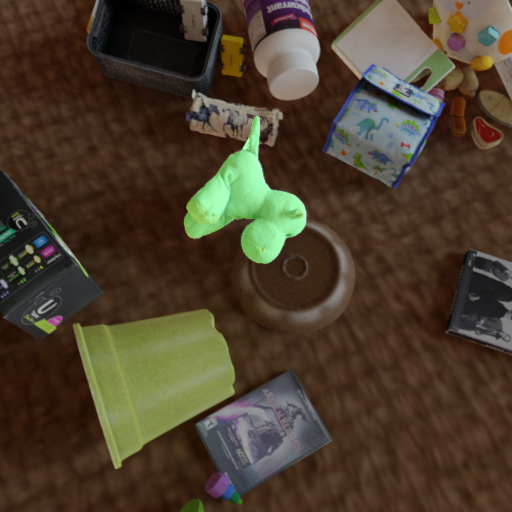}}\hspace{\liftwidth}
	
    \caption{\textbf{Qualitative Comparison with Panoptic Lifting and Contrastive Lift}. Zoom in for details.}
    \label{fig:comparison_lift}
\end{figure}

\newcommand\rgbWidth{0.01cm}
\newcommand\rgbHeight{0.05cm}

\begin{figure}[t]
    \centering
    \captionsetup[subfloat]{position=top, labelformat=empty}
    
    \subfloat[Ground Truth]{\includegraphics[width = 0.32\linewidth]{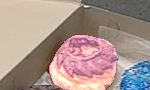}}\hspace{\rgbWidth}
    \subfloat[w/o $\mathcal{L}_{RGB}$ ]{\includegraphics[width = 0.32\linewidth]{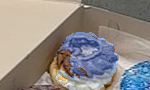}}\hspace{\rgbWidth}
    \subfloat[w/ $\mathcal{L}_{RGB}$ ]{\includegraphics[width = 0.32\linewidth]{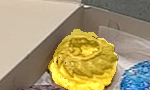}}\hspace{\rgbWidth}

    \vspace{\rgbHeight}
    
    \subfloat{\includegraphics[width = 0.32\linewidth]{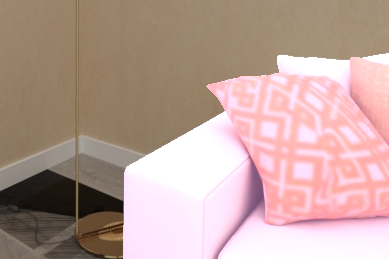}}\hspace{\rgbWidth}
    \subfloat{\includegraphics[width = 0.32\linewidth]{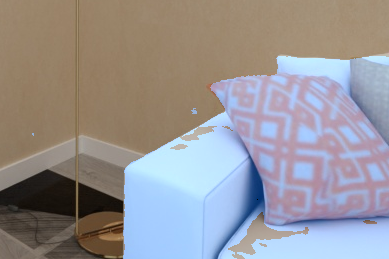}}\hspace{\rgbWidth}
    \subfloat{\includegraphics[width = 0.32\linewidth]{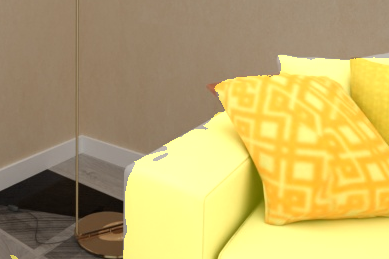}}\hspace{\rgbWidth}

    \vspace{-0.1in}
    \caption{ \textbf{Qualitative Results of the Ray-Pair RGB Loss.} The Ray-Pair RGB loss can help to recover local regions and make the results more solid.}
    \label{fig:ablation_rgb}
\end{figure}

\vspace{0.5cm}
\section{Extending to Dynamic NeRFs}
We present a preliminary demonstration in Figure~\ref{fig:4dnerf} on the easy extension of our method to 4D dynamic NeRF representations. We use HyperReel~\cite{attal2023hyperreel} as our reference NeRF representation and only supply user prompts for the first frame of each camera. The prompts are fed into SAM to retrieve initial masks, whose bounding boxes are used as the prompts for the next frame. This process repeats until masks are acquired from all video frames, after which we proceed to object field training as in previous static scene cases. The scene is from the Neural 3D Video dataset~\cite{li2022neural}.

\newcommand\dynamicNeRFWidth{0.02cm}
\newcommand\dynamicNeRFHeight{0.02cm}

\begin{figure*}[t]
    \centering
    \captionsetup[subfloat]{position=bottom}

    {\includegraphics[width = 0.13\linewidth]{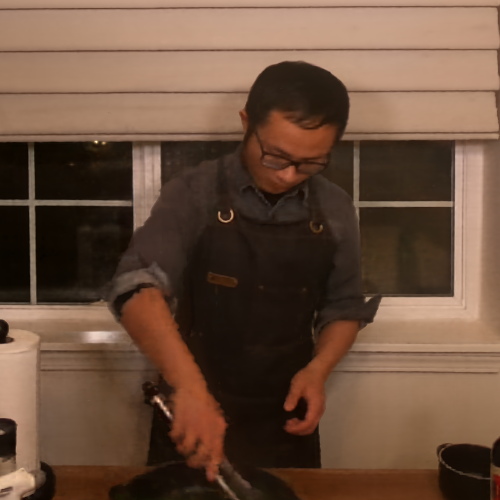}}\hspace{\dynamicNeRFWidth}
    {\includegraphics[width = 0.13\linewidth]{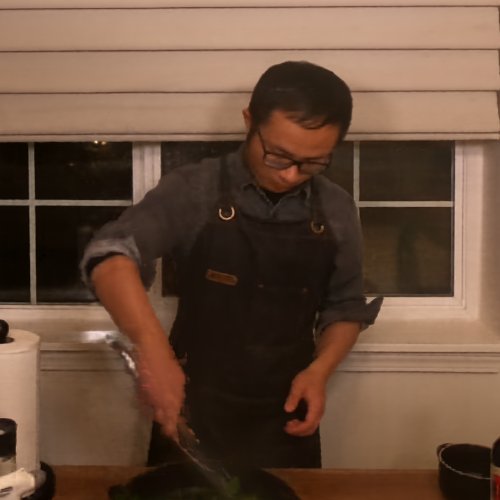}}\hspace{\dynamicNeRFWidth}
    {\includegraphics[width = 0.13\linewidth]{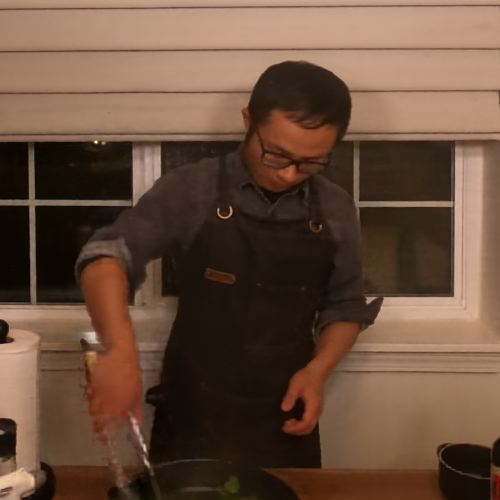}}\hspace{\dynamicNeRFWidth}
    {\includegraphics[width = 0.13\linewidth]{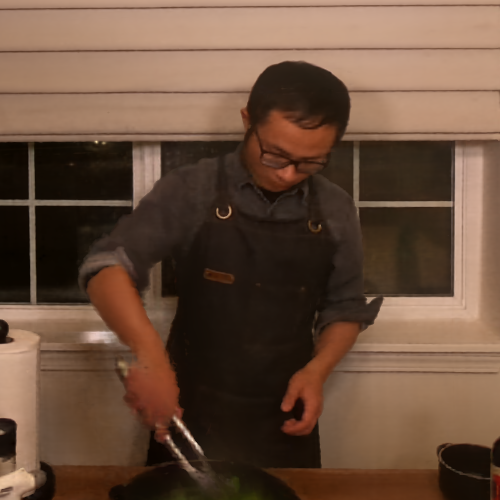}}\hspace{\dynamicNeRFWidth}   
    {\includegraphics[width = 0.13\linewidth]{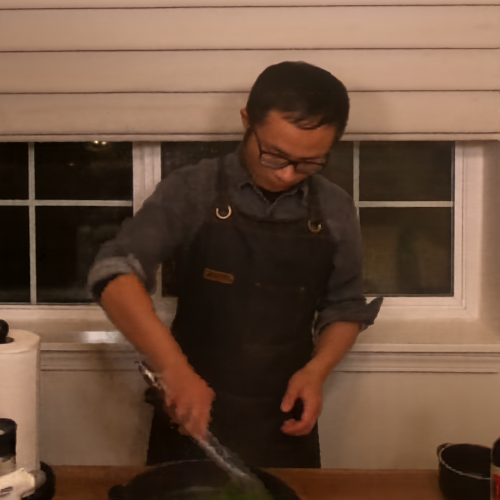}}\hspace{\dynamicNeRFWidth}
    {\includegraphics[width = 0.13\linewidth]{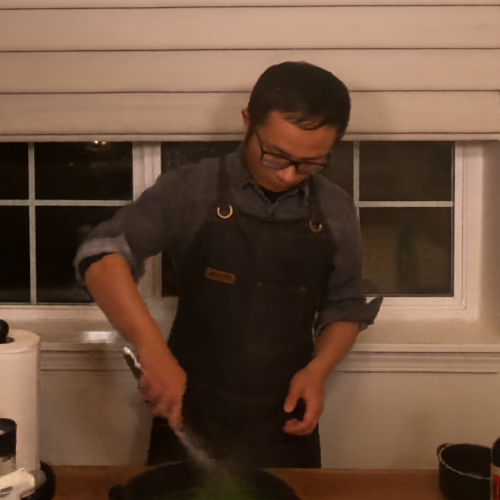}}\hspace{\dynamicNeRFWidth}   
    {\includegraphics[width = 0.13\linewidth]{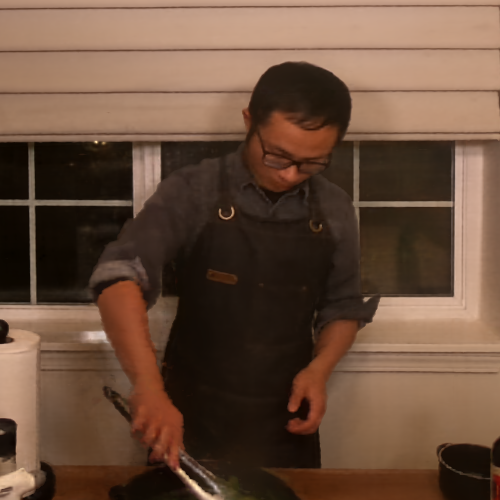}}

    \vspace{\dynamicNeRFHeight}

    \subfloat[$t = 0s$]{\includegraphics[width = 0.13\linewidth]{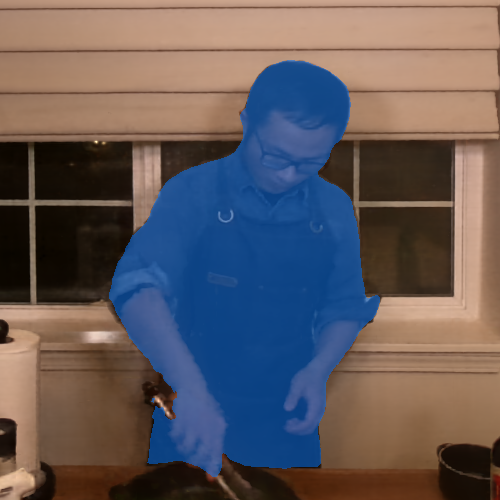}}\hspace{\dynamicNeRFWidth}
    \subfloat[$t = 0.25s$]{\includegraphics[width = 0.13\linewidth]{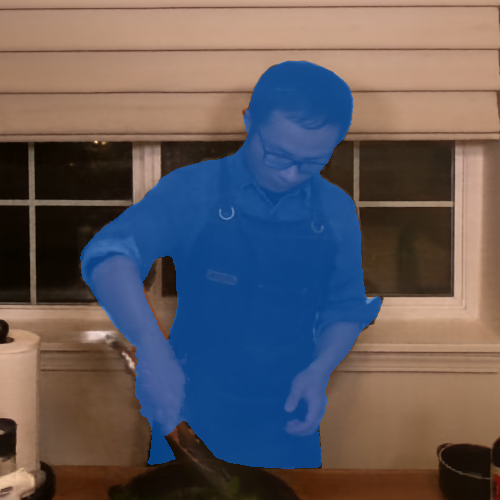}}\hspace{\dynamicNeRFWidth}
    \subfloat[$t = 0.5s$]{\includegraphics[width = 0.13\linewidth]{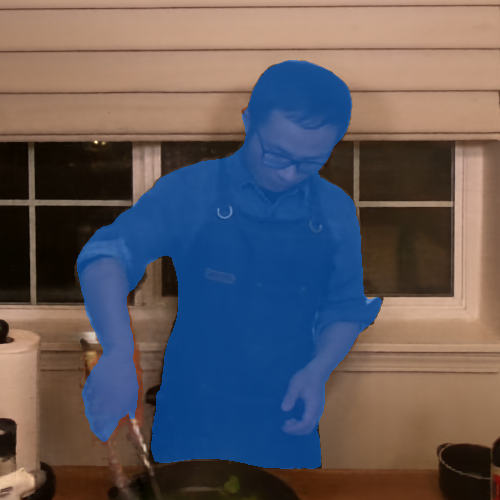}}\hspace{\dynamicNeRFWidth}
    \subfloat[$t = 0.75s$]{\includegraphics[width = 0.13\linewidth]{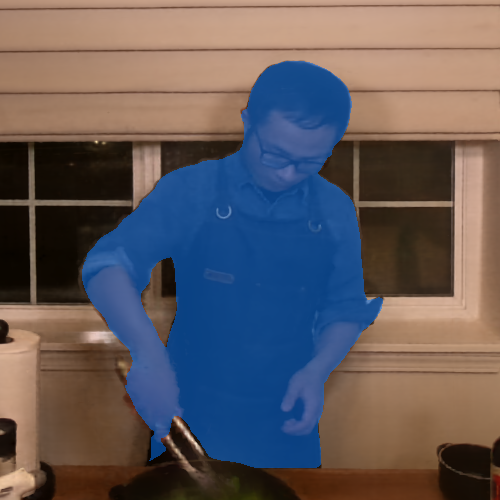}}\hspace{\dynamicNeRFWidth}
    \subfloat[$t = 1s$]{\includegraphics[width = 0.13\linewidth]{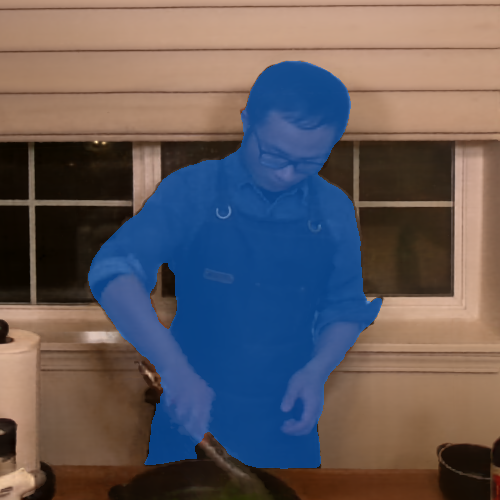}}\hspace{\dynamicNeRFWidth}
    \subfloat[$t = 1.25s$]{\includegraphics[width = 0.13\linewidth]{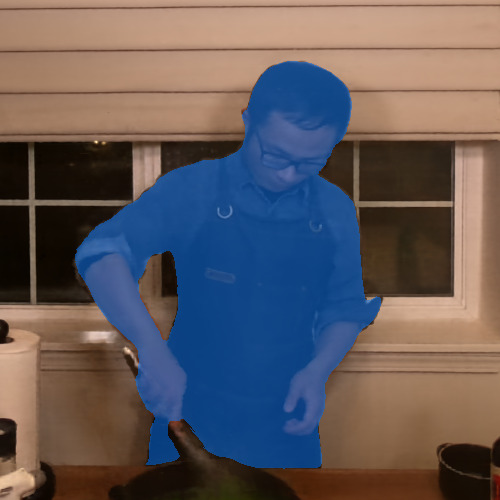}}\hspace{\dynamicNeRFWidth}
    \subfloat[$t = 1.5s$]{\includegraphics[width = 0.13\linewidth]{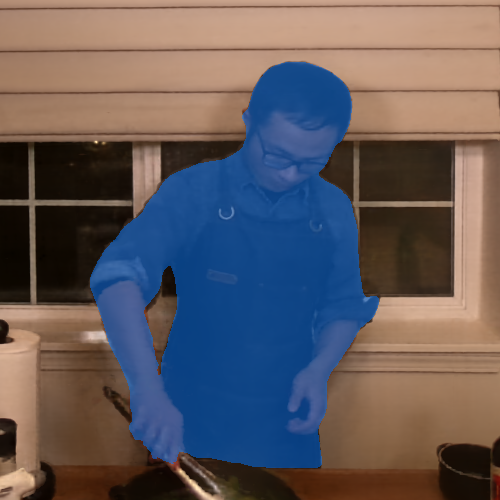}}

    \vspace{-0.05in}
    \caption{\textbf{Demonstration of Applying SANeRF-HQ to Dynamic NeRFs.} The first row are the NeRF RGB images over time, and the second row are the masks from SANeRF-HQ, which is also dynamic. Our method can be easily adapted to dynamic NeRFs and still retains reasonable performance. The implementation is based on HyperReel, and the \textit{cook spinach} scene shown is from the Neural 3D Video dataset.}
    \label{fig:4dnerf}
\end{figure*}

\newcommand\replicaWidth{0.02cm}
\newcommand\replicaHeight{0.02cm}

\begin{figure*}[t]
    \centering
    \captionsetup[subfloat]{position=top}

    \subfloat[RGB]{\includegraphics[width = 0.18\linewidth]{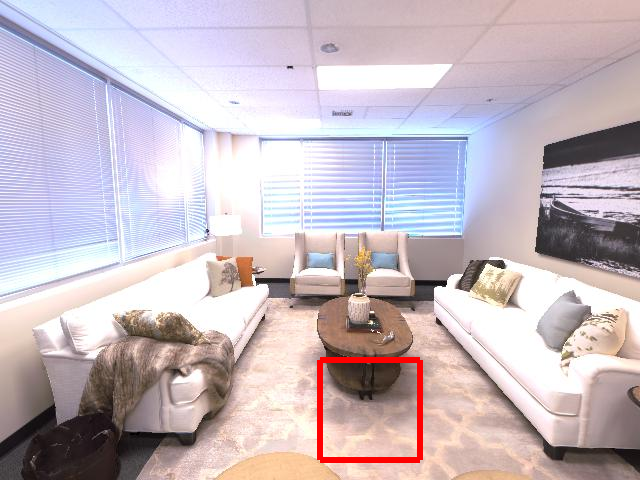}}\hspace{\replicaWidth}
    \subfloat[GT]{\includegraphics[width = 0.18\linewidth]{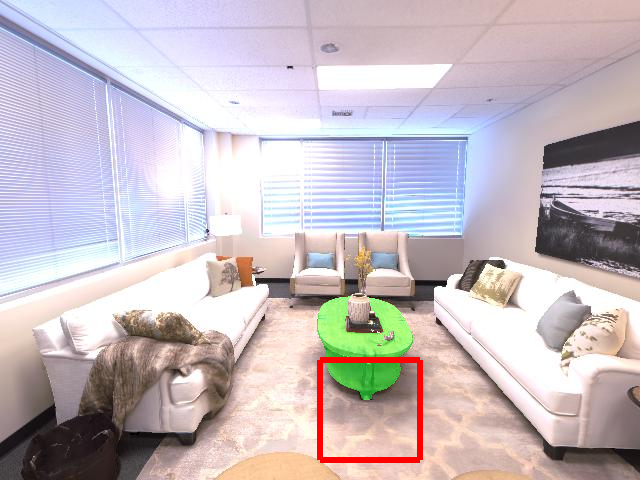}}\hspace{\replicaWidth}
    \subfloat[ISRF]{\includegraphics[width = 0.18\linewidth]{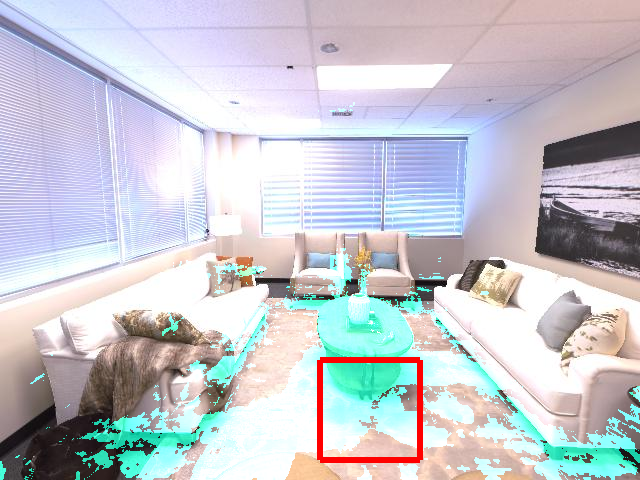}}\hspace{\replicaWidth}
    \subfloat[SA3D]{\includegraphics[width = 0.18\linewidth]{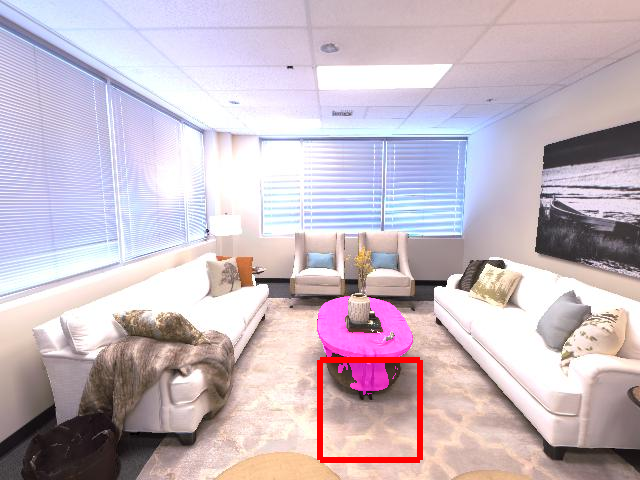}}\hspace{\replicaWidth}   
    \subfloat[Ours]{\includegraphics[width = 0.18\linewidth]{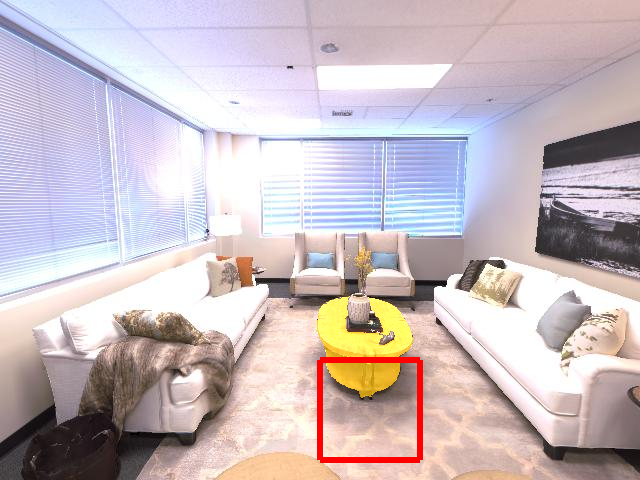}}\hspace{\replicaWidth}

    \vspace{\replicaHeight}    
    \subfloat{\includegraphics[width = 0.18\linewidth]{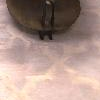}}\hspace{\replicaWidth}
    \subfloat{\includegraphics[width = 0.18\linewidth]{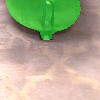}}\hspace{\replicaWidth}
    \subfloat{\includegraphics[width = 0.18\linewidth]{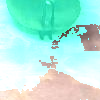}}\hspace{\replicaWidth}
    \subfloat{\includegraphics[width = 0.18\linewidth]{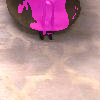}}\hspace{\replicaWidth}  
    \subfloat{\includegraphics[width = 0.18\linewidth]{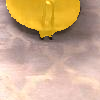}}\hspace{\replicaWidth}

\vspace{-0.05in}
\caption{\textbf{Comparison with SA3D and ISRF on the Replica Room.} Data is from the Others subset. SANeRF-HQ can maintain the object structure while excludes the background. \vspace{-0.0in}}
\label{fig:comparison_3d_replica}
\end{figure*}

\newcommand\shoerackWidth{0.02cm}
\newcommand\shoerackHeight{0.02cm}

\begin{figure*}[h]
    \centering
    \captionsetup[subfloat]{position=top}
    \subfloat[RGB]{\includegraphics[width = 0.18\linewidth]{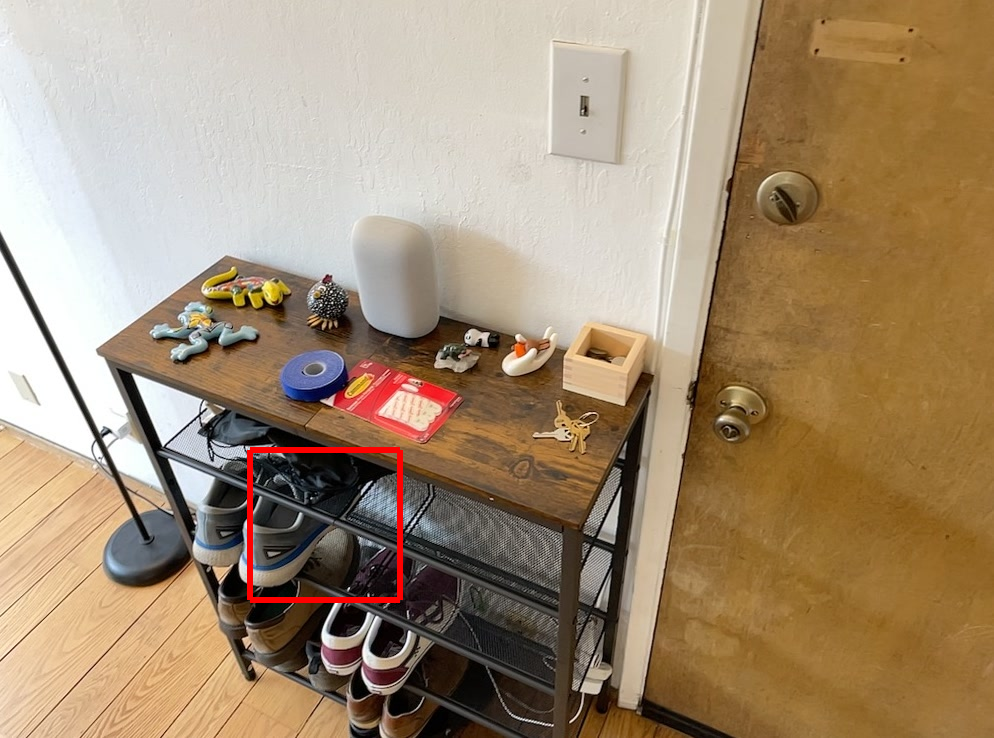}}\hspace{\shoerackWidth}
    \subfloat[GT]{\includegraphics[width = 0.18\linewidth]{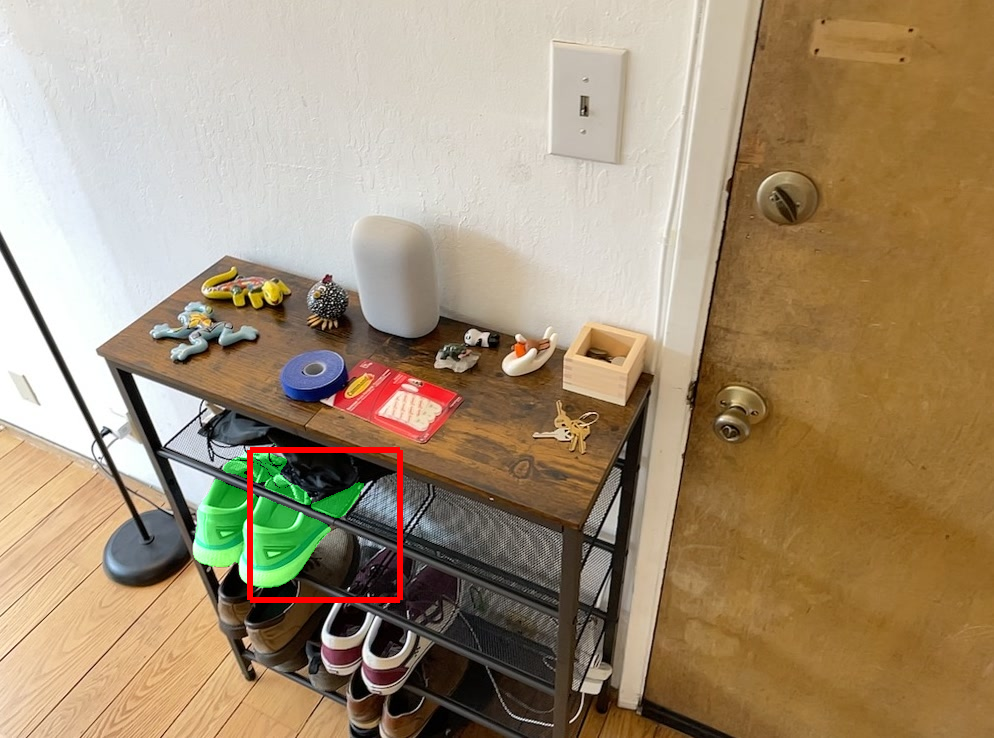}}\hspace{\shoerackWidth}
    \subfloat[ISRF]{\includegraphics[width = 0.18\linewidth]{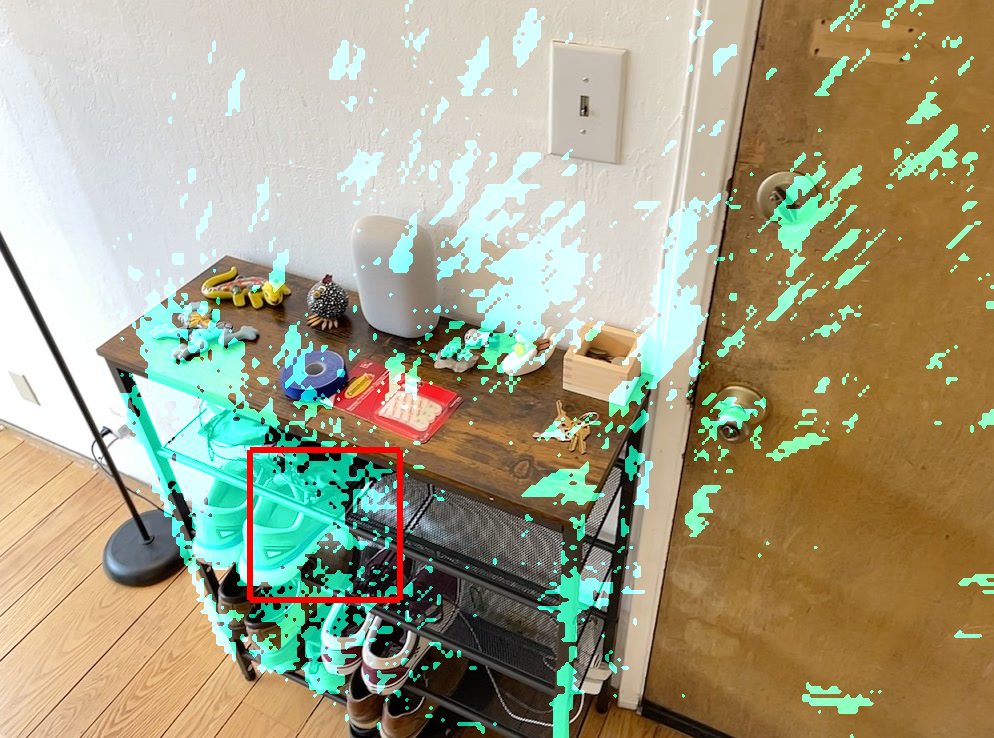}}\hspace{\shoerackWidth}
    \subfloat[SA3D]{\includegraphics[width = 0.18\linewidth]{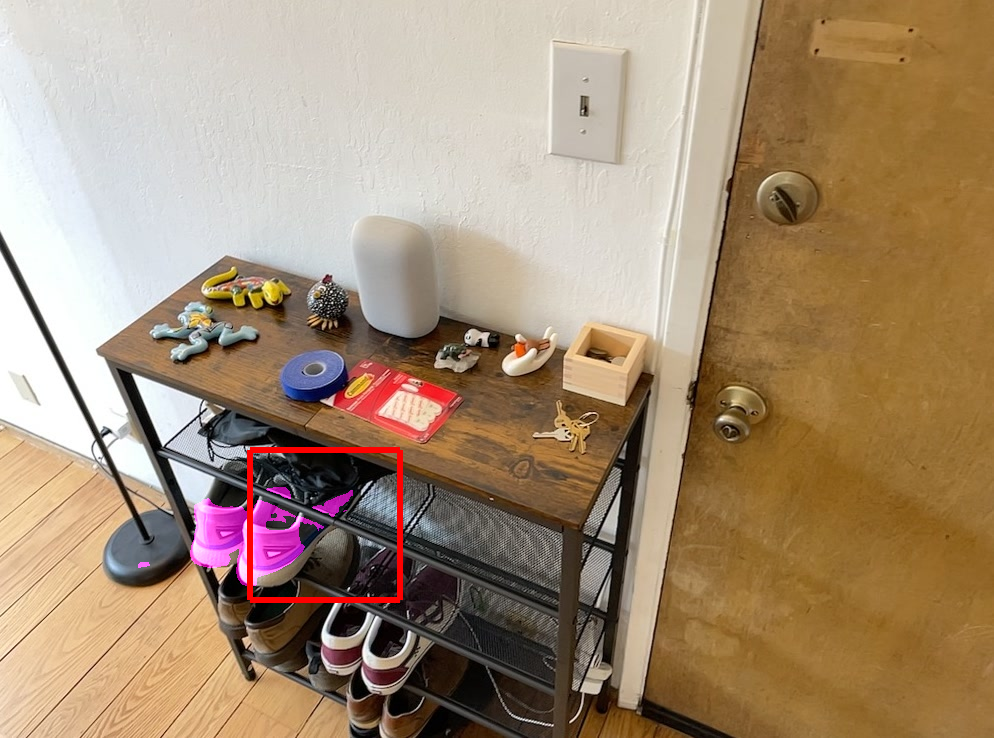}}\hspace{\shoerackWidth}   
    \subfloat[Ours]{\includegraphics[width = 0.18\linewidth]{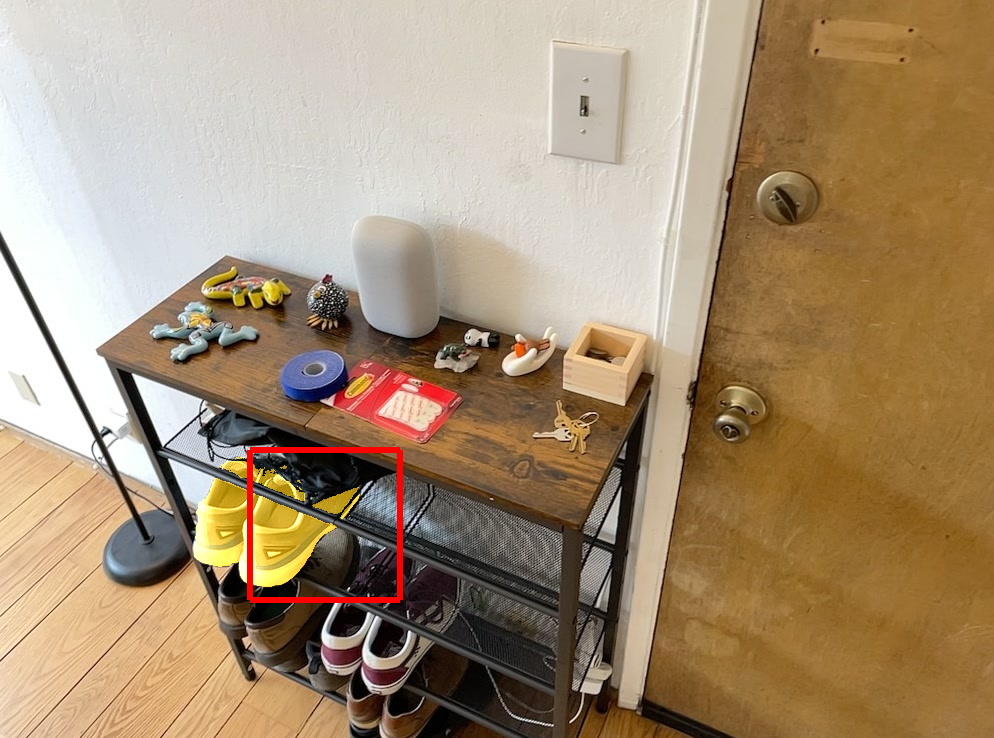}}\hspace{\shoerackWidth}

    \vspace{\shoerackHeight}    
    \subfloat{\includegraphics[width = 0.18\linewidth]{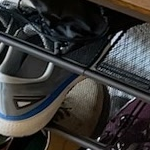}}\hspace{\shoerackWidth}
    \subfloat{\includegraphics[width = 0.18\linewidth]{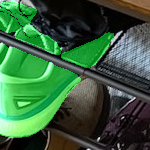}}\hspace{\shoerackWidth}
    \subfloat{\includegraphics[width = 0.18\linewidth]{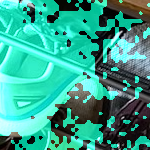}}\hspace{\shoerackWidth}
    \subfloat{\includegraphics[width = 0.18\linewidth]{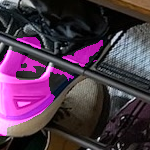}}\hspace{\shoerackWidth}  
    \subfloat{\includegraphics[width = 0.18\linewidth]{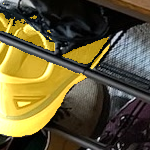}}\hspace{\shoerackWidth}

\vspace{-0.05in}
\caption{\textbf{Comparison with SA3D and ISRF on the Shoe Rack.} Data is from the LERF subset. Our method can reproduce the segmentation details even with some occlusion. \vspace{-0.0in}}

\label{fig:comparison_3d_shoe_rack}
\end{figure*}

\newcommand\aiWidth{0.02cm}
\newcommand\aiHeight{0.02cm}

\begin{figure*}[h]

    \centering
    \captionsetup[subfloat]{position=top}

    \subfloat[RGB]{\includegraphics[width = 0.18\linewidth]{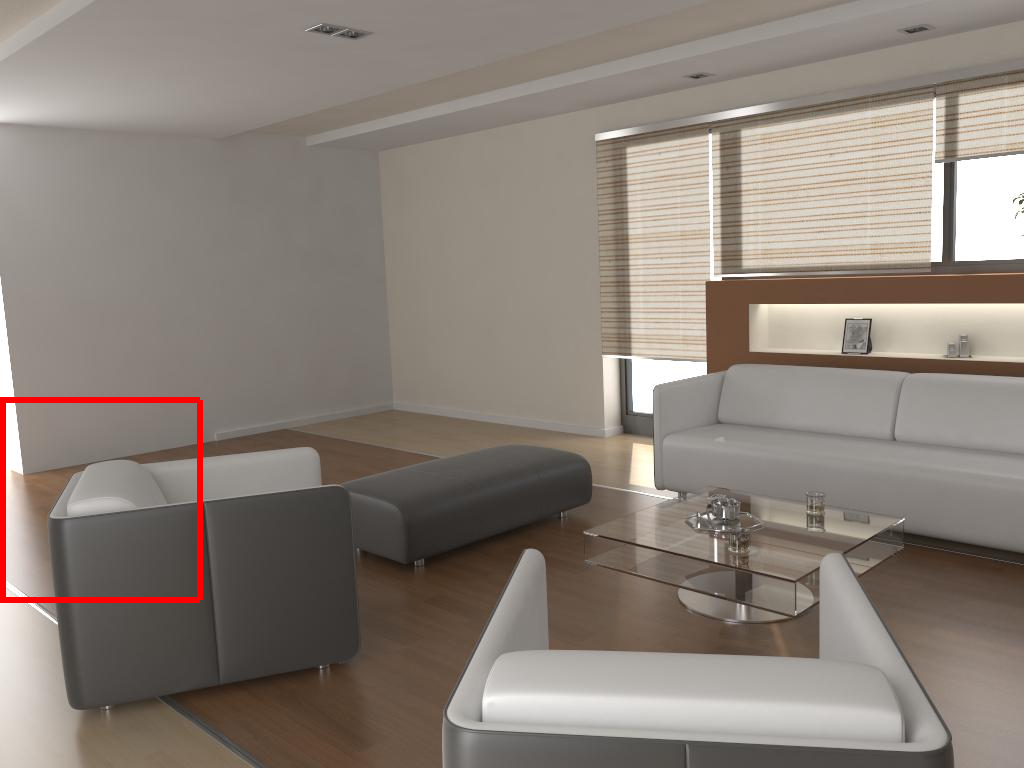}}\hspace{\aiWidth}
    \subfloat[GT]{\includegraphics[width = 0.18\linewidth]{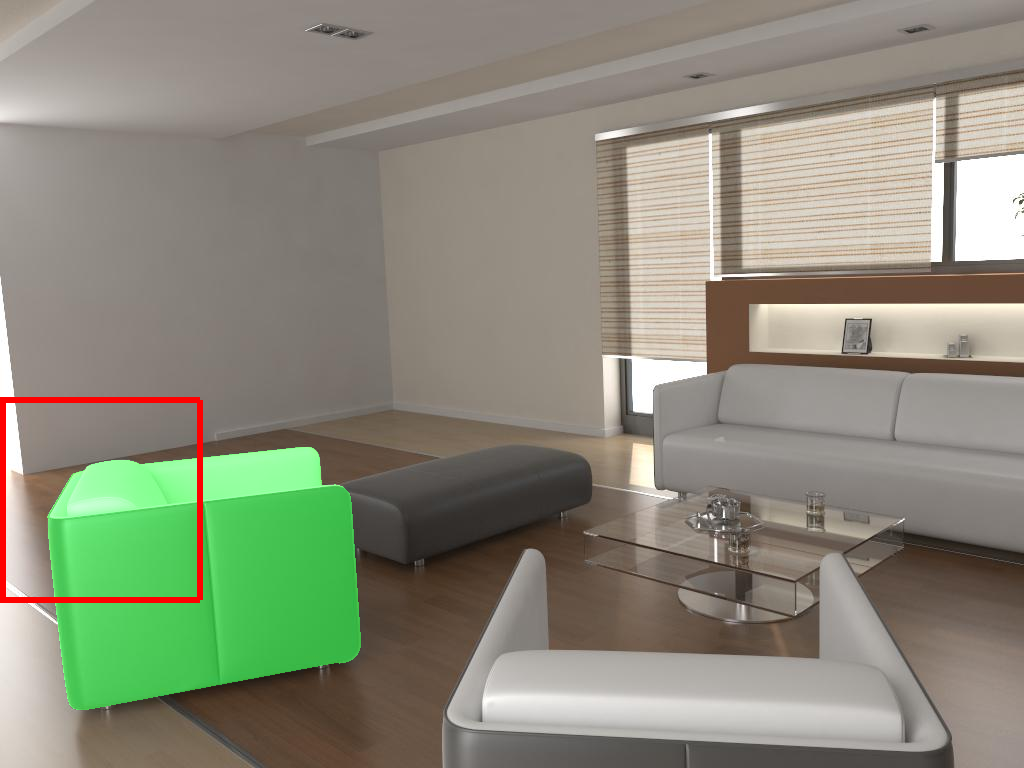}}\hspace{\aiWidth}
    \subfloat[ISRF]{\includegraphics[width = 0.18\linewidth]{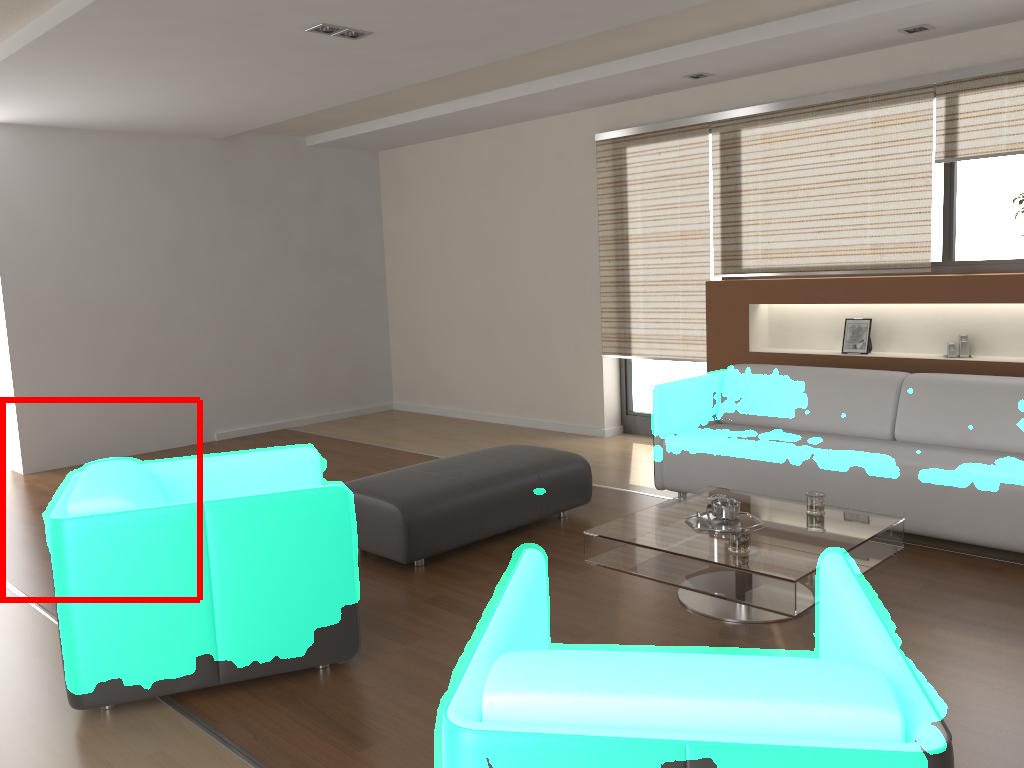}}\hspace{\aiWidth}
    \subfloat[SA3D]{\includegraphics[width = 0.18\linewidth]{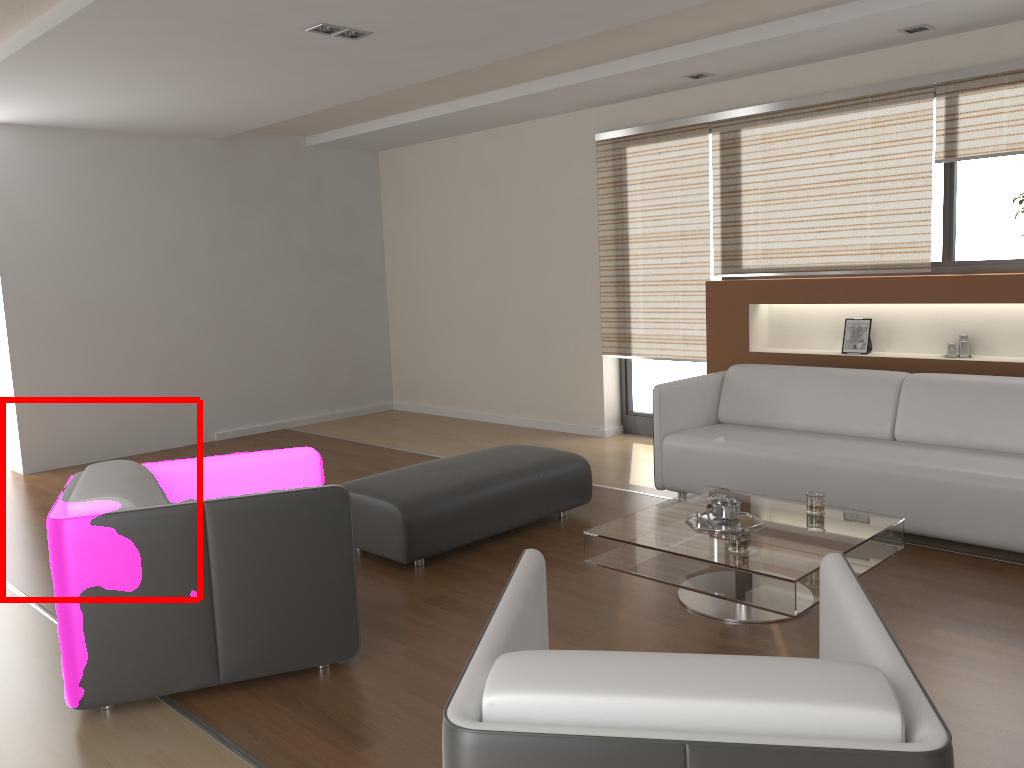}}\hspace{\aiWidth}   
    \subfloat[Ours]{\includegraphics[width = 0.18\linewidth]{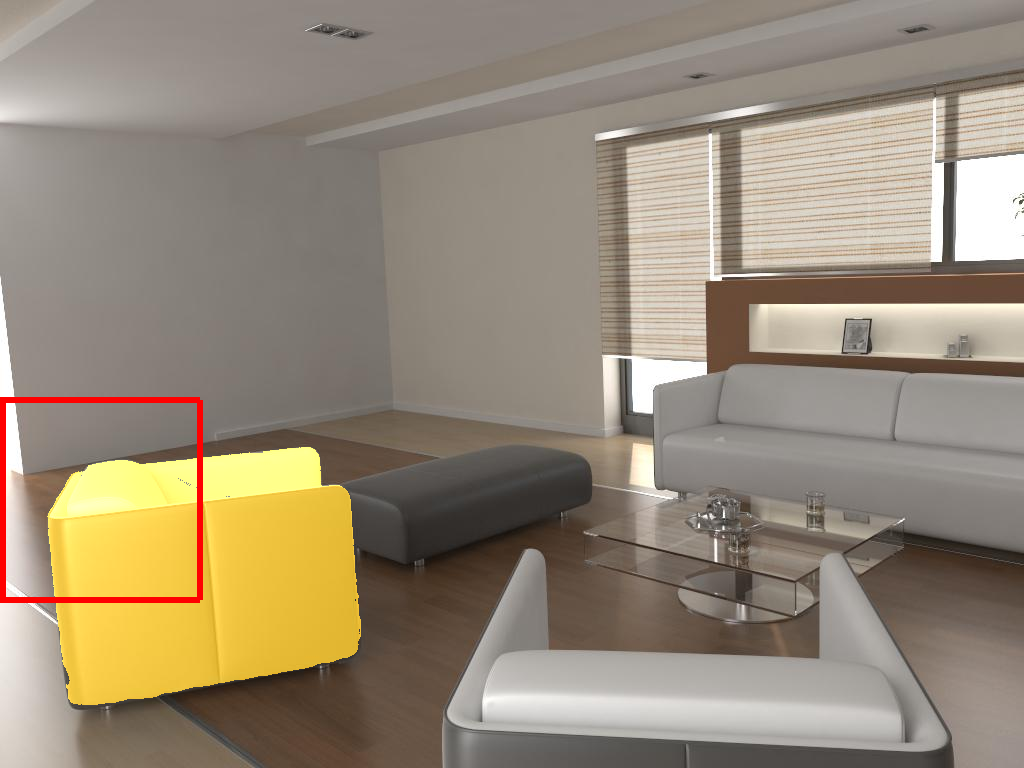}}\hspace{\aiWidth}

    \vspace{\aiHeight}    
    \subfloat{\includegraphics[width = 0.18\linewidth]{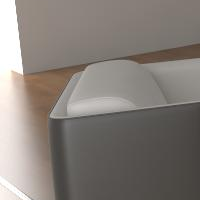}}\hspace{\aiWidth}
    \subfloat{\includegraphics[width = 0.18\linewidth]{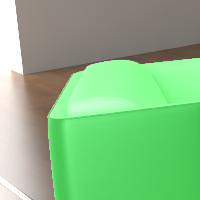}}\hspace{\aiWidth}
    \subfloat{\includegraphics[width = 0.18\linewidth]{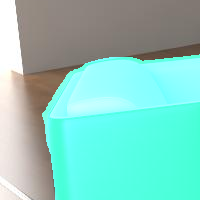}}\hspace{\aiWidth}
    \subfloat{\includegraphics[width = 0.18\linewidth]{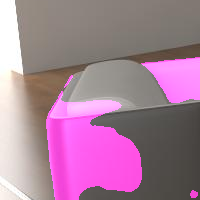}}\hspace{\aiWidth}  
    \subfloat{\includegraphics[width = 0.18\linewidth]{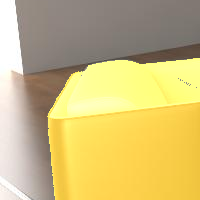}}\hspace{\aiWidth}

\vspace{-0.05in}
\caption{\textbf{Comparison with SA3D and ISRF on Hypersim.} Data is from the Others subset. ISRF contains too many false positives, while SA3D cannot cover the whole object. \vspace{-0.1in}}
    
\label{fig:comparison_3d_ai_001_008}
\end{figure*}

\newcommand\espressoWidth{0.02cm}
\newcommand\espressoHeight{0.02cm}

\begin{figure*}[ht]

    \centering
    \captionsetup[subfloat]{position=top}

    \subfloat[RGB]{\includegraphics[width = 0.18\linewidth]{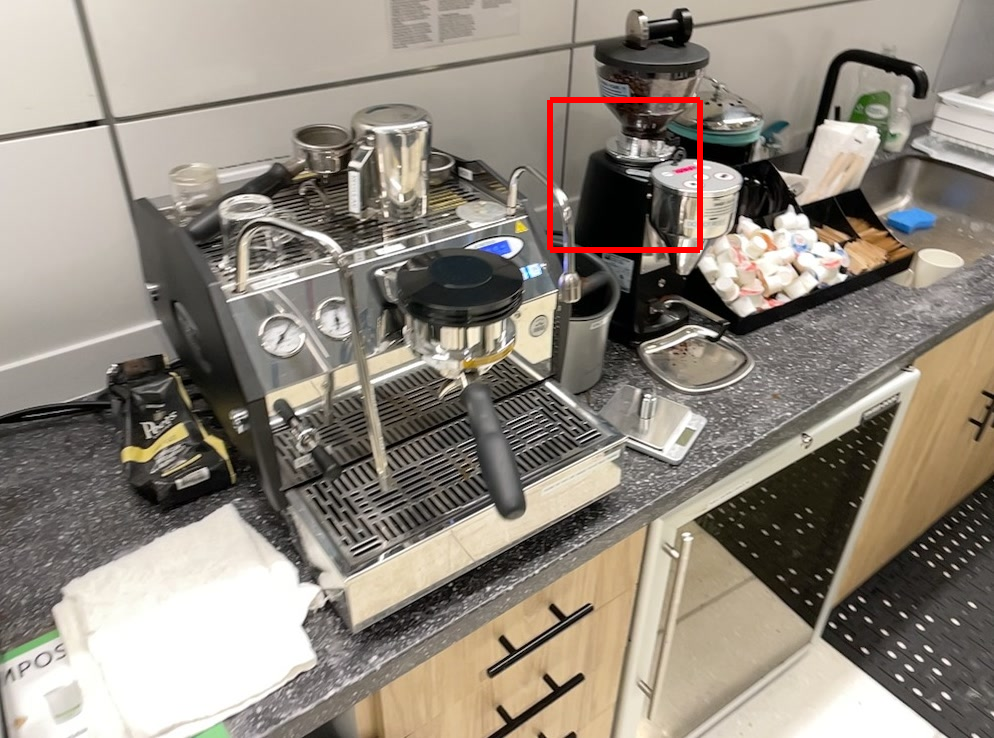}}\hspace{\espressoWidth}
    \subfloat[GT]{\includegraphics[width = 0.18\linewidth]{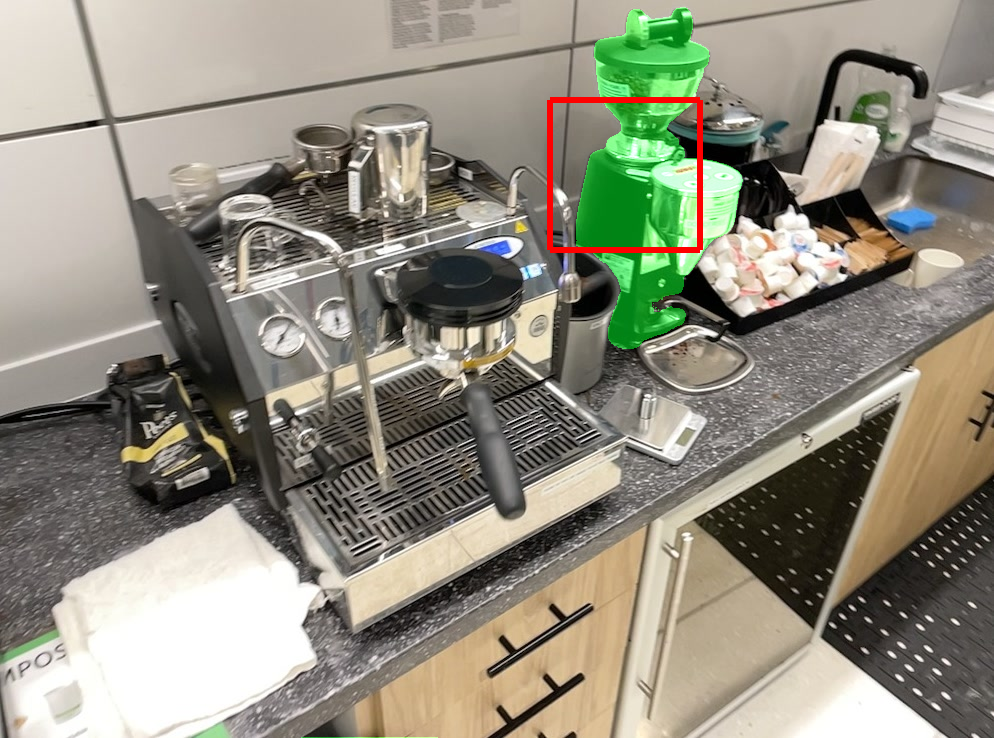}}\hspace{\espressoWidth}
    \subfloat[ISRF]{\includegraphics[width = 0.18\linewidth]{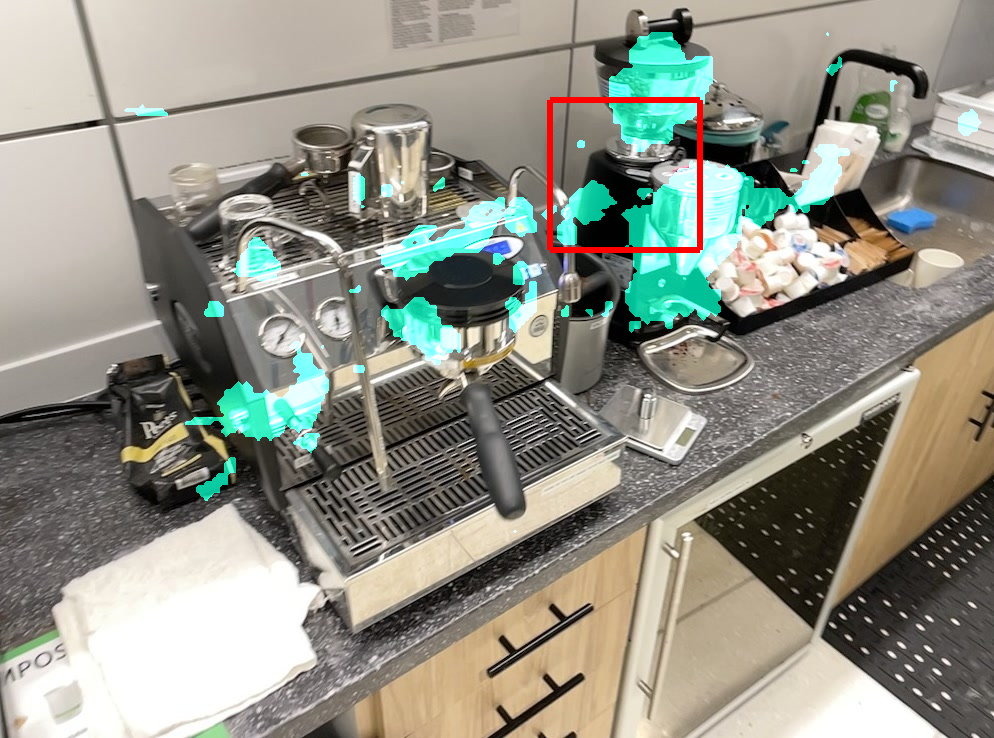}}\hspace{\espressoWidth}
    \subfloat[SA3D]{\includegraphics[width = 0.18\linewidth]{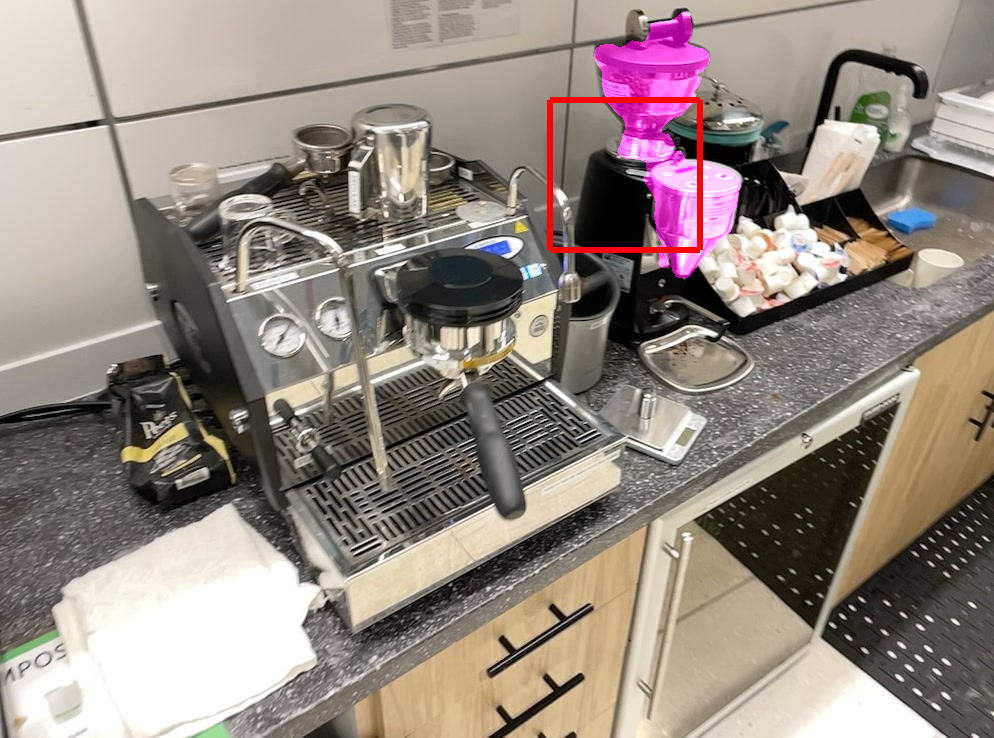}}\hspace{\espressoWidth}   
    \subfloat[Ours]{\includegraphics[width = 0.18\linewidth]{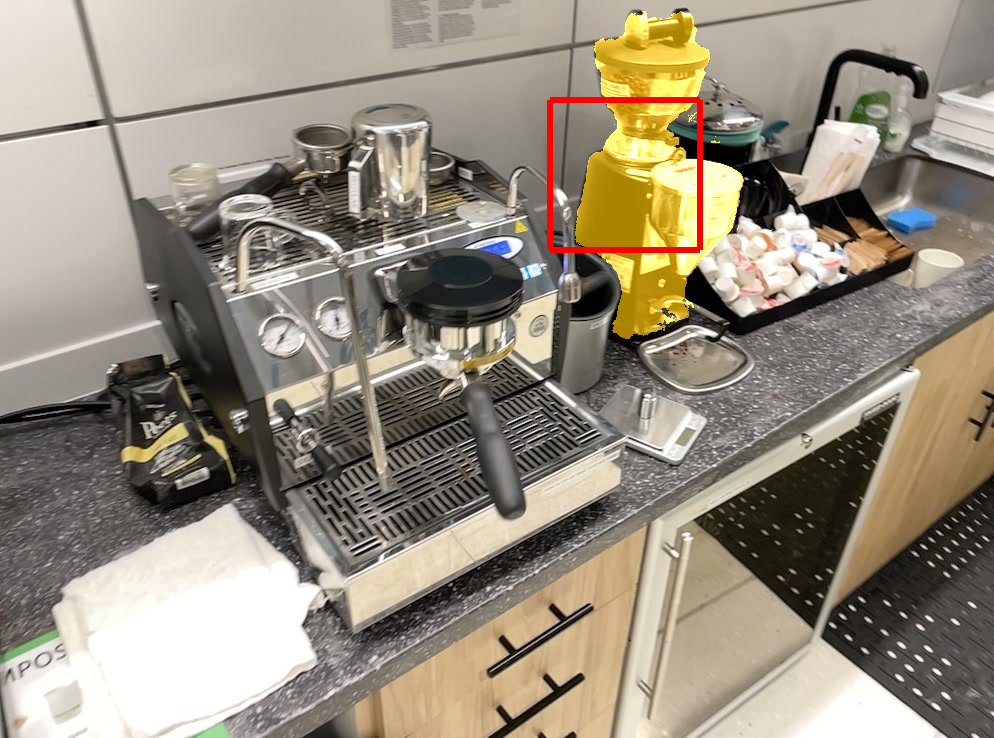}}\hspace{\espressoWidth}

    \vspace{\espressoHeight}    
    \subfloat{\includegraphics[width = 0.18\linewidth]{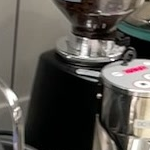}}\hspace{\espressoWidth}
    \subfloat{\includegraphics[width = 0.18\linewidth]{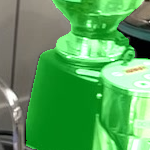}}\hspace{\espressoWidth}
    \subfloat{\includegraphics[width = 0.18\linewidth]{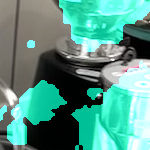}}\hspace{\espressoWidth}
    \subfloat{\includegraphics[width = 0.18\linewidth]{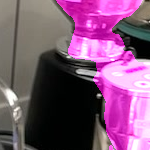}}\hspace{\espressoWidth}  
    \subfloat{\includegraphics[width = 0.18\linewidth]{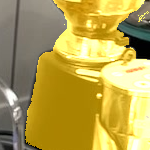}}\hspace{\espressoWidth}

\vspace{-0.05in}
\caption{\textbf{Comparison with SA3D and ISRF on the Espresso.} Data is from the LERF subset. Our method produces the most reasonable segmentation in the distant, complex setting. \vspace{-0.1in}}
    
\label{fig:comparison_3d_espresso}
\end{figure*}

\section{More Qualitative Results}
We demonstrate the qualitative results of the Ray-Pair RGB loss in Figure~\ref{fig:ablation_rgb}. The loss helps fill in the missing interior and boundaries of the masks by enforcing a local match between the similarity in labels, and the similarity in appearance. 

We also provide extra qualitative comparisons between our method and other zero-shot 3D segmentation methods mentioned in the main paper. The results are given in Figures~\ref{fig:comparison_3d_replica},~\ref{fig:comparison_3d_shoe_rack},~\ref{fig:comparison_3d_ai_001_008},~\ref{fig:comparison_3d_espresso}. Please watch the video  for more qualitative results.

\section{Limitations}
Though our method works well in most cases, it relies on NeRF and SAM, and its performance might be impacted by scene complexity and NeRF quality. On the other hand, the Ray-Pair RGB loss may not handle all circumstances especially given neighboring objects with identical colors and shading. Nevertheless, we present some results of our method on relatively challenging scenes to show that it may still robustly handle some of these cases, where the target objects are relatively small, in the background, partially occluded, or adjacent to other objects with similar appearance. The results are in Figure \ref{fig:occlusion} and \ref{fig:color}. We leave relevant potential improvements as future work.

\newcommand{\colorWidth}{0.01cm}
\begin{figure*}[ht]
    \centering
    \captionsetup[subfloat]{position=top, labelformat=empty}
    
    \subfloat[RGB]{\includegraphics[width =0.24\linewidth]{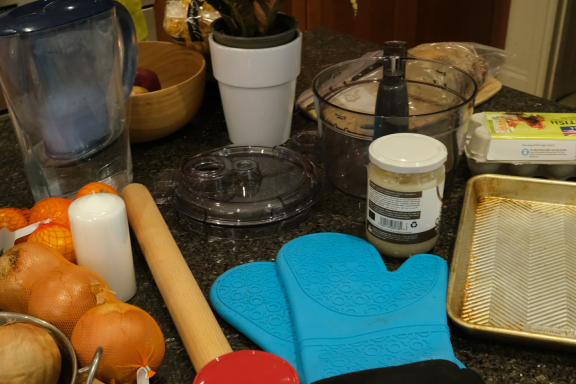}}\hspace{\colorWidth}
    \subfloat[Ours]{\includegraphics[width =0.24\linewidth]{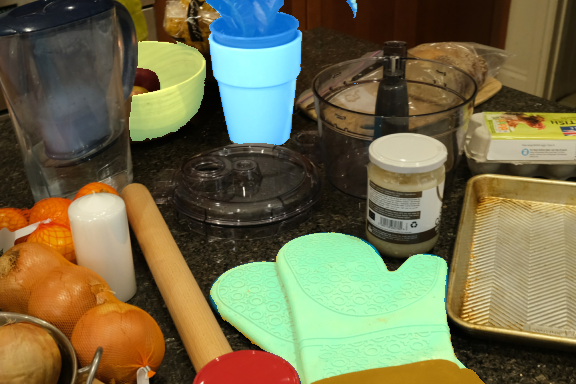}}\hspace{\colorWidth}
    \subfloat[RGB]{\includegraphics[width =0.24\linewidth]{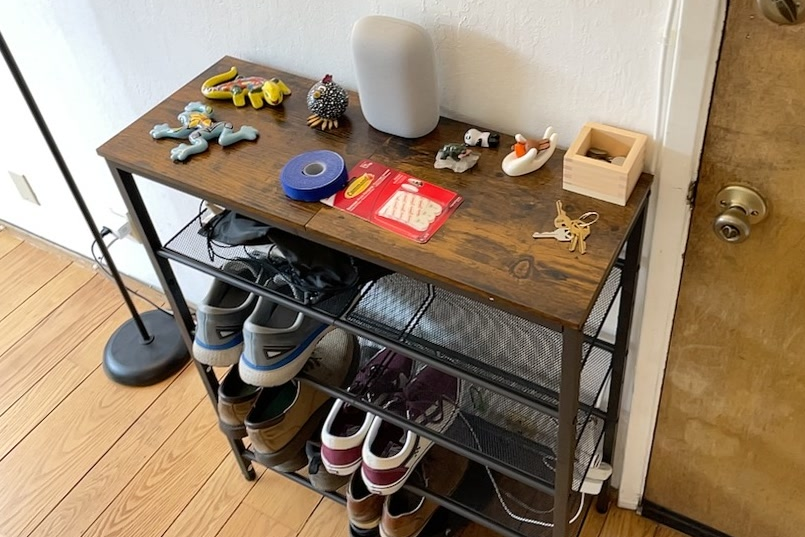}}\hspace{\colorWidth}
    \subfloat[Ours]{\includegraphics[width =0.24\linewidth]{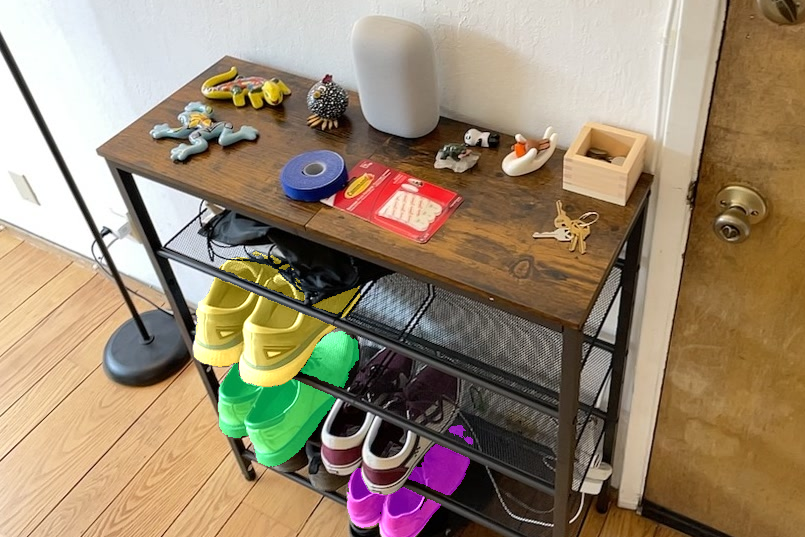}}\hspace{\colorWidth}

    \vspace{-0.05in}
    \caption{\textbf{ Examples of More Complex Scenarios. } SANeRF-HQ can effectively segment target objects that are in the background, relatively small, and partially occluded.}
    \label{fig:occlusion}
\end{figure*}

\renewcommand\colorWidth{0.0cm}
\begin{figure*}[ht]
    \centering
    \captionsetup[subfloat]{position=top, labelformat=empty}
    
    \subfloat{\includegraphics[width =0.19\linewidth]{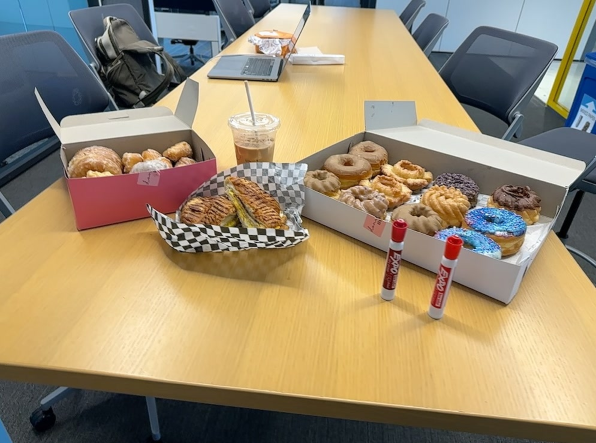}}\hspace{\colorWidth}
    \subfloat{\includegraphics[width =0.19\linewidth]{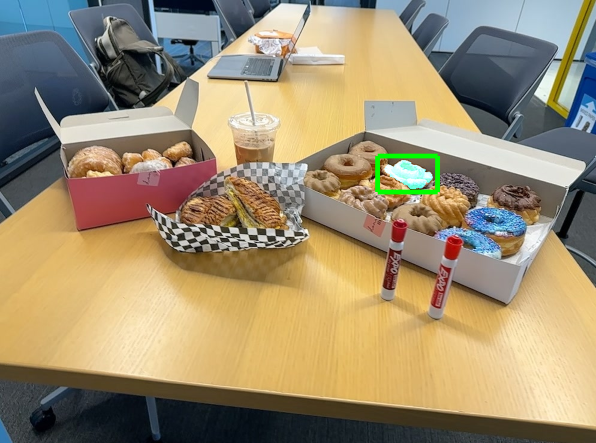}}\hspace{\colorWidth}
    \subfloat{\includegraphics[width =0.19\linewidth]{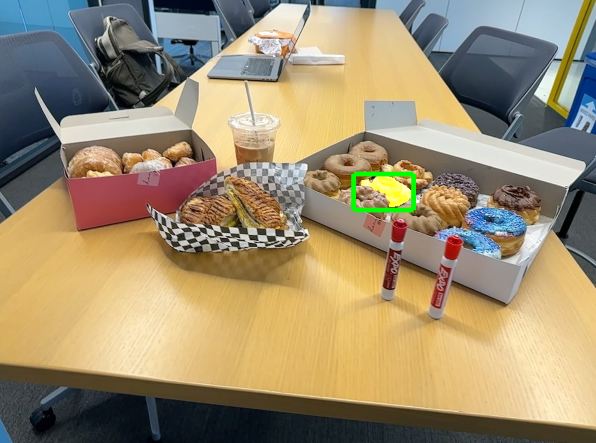}}\hspace{\colorWidth}
    \subfloat{\includegraphics[width =0.19\linewidth]{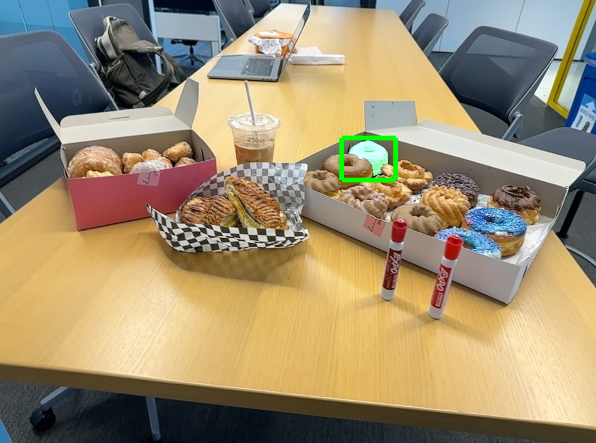}}\hspace{\colorWidth}
    \subfloat{\includegraphics[width =0.19\linewidth]{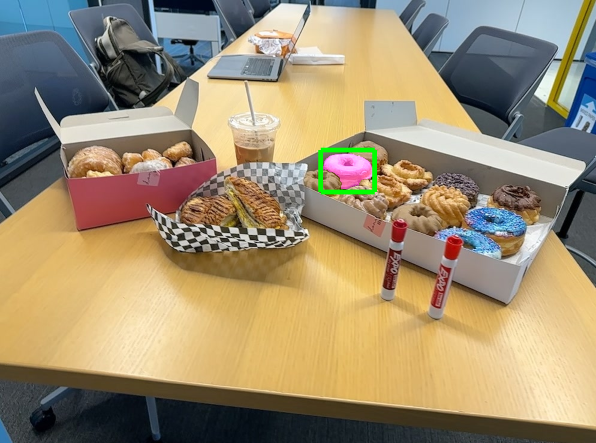}}\hspace{\colorWidth}

    \vspace{-0.05in}
    \caption{\textbf{ Examples of Objects with Similar Color. } Our method can still distinguish these objects and produce reasonable results in the presence of neighbouring objects with similar appearance, where the Ray-Pair RGB loss is less helpful but remains robust. }
    \label{fig:color}
\end{figure*}

\end{document}